\documentclass[sigconf]{acmart}

\renewcommand\footnotetextcopyrightpermission[1]{}
\setcopyright{none}

\AtBeginDocument{%
  }
\usepackage{algorithm}
\usepackage{enumitem}
\usepackage{multirow}
\usepackage{amsmath}
\usepackage{algorithmic}
\usepackage{listings}
\usepackage{subfigure}
\usepackage{graphicx}
\usepackage{blindtext}
\usepackage{array}
\usepackage{dutchcal}

\newcommand{\ie}{\emph{i.e.}\xspace} 
\newcommand{\etc}{\emph{etc.}\xspace} 

\newcommand{\our}{{AnomalyLLM}\xspace}

\begin{document}
\title{\our: Few-shot Anomaly Edge Detection for Dynamic Graphs using Large Language Models}

\author{Shuo Liu}
\affiliation{
    \institution{
    Institute of Computing Technology, Chinese Academy of Sciences\\
    }
    \country{} 
}

\author{Di Yao}
\affiliation{
    \institution{
    Institute of Computing Technology, Chinese Academy of Sciences\\
    }
    \country{} 
}
\email{yaodi@ict.ac.cn}
\authornote{Corresponding authors.}

\author{Lanting Fang}
\affiliation{
    \institution{
    	Beijing Institute of Technology\\
    }
    \country{} 
}

\author{Zhetao Li}
\affiliation{
    \institution{
      Jinan University\\
    }
    \country{} 
}

\author{Wenbin Li}
\affiliation{
    \institution{
      Institute of Computing Technology, Chinese Academy of Sciences\\
    }
    \country{} 
}
\author{Kaiyu Feng}
\affiliation{
    \institution{
    	Beijing Institute of Technology\\
    }
    \country{} 
}

\author{Xiaowen Ji}
\affiliation{
    \institution{
      Southeast University\\
    }
    \country{} 
}
\author{Jingping Bi}
\affiliation{
    \institution{
    Institute of Computing Technology, Chinese Academy of Sciences\\
    }
    \country{} 
}
\email{bjp@ict.ac.cn}
\authornotemark[1]

\begin{abstract}
Detecting anomaly edges for dynamic graphs aims to identify edges significantly deviating from the normal pattern and can be applied in various domains, such as cybersecurity, financial transactions and AIOps. With the evolving of time, the types of anomaly edges are emerging and the labeled anomaly samples are few for each type. Current methods are either designed to detect randomly inserted edges or require sufficient labeled data for model training, which harms their applicability for real-world applications. In this paper, we study this problem by cooperating with the rich knowledge encoded in large language models(LLMs) and propose a method, namely \our. To align the dynamic graph with LLMs, \our pre-trains a dynamic-aware encoder to generate the representations of edges and reprograms the edges using the prototypes of word embeddings. Along with the encoder, we design an in-context learning framework that integrates the information of a few labeled samples to achieve few-shot anomaly detection. Experiments on four datasets reveal that \our can not only significantly improve the performance of few-shot anomaly detection, but also achieve superior results on new anomalies without any update of model parameters.
\end{abstract}
  
\keywords{Dynamic Graphs, Anomaly Detection, Few-Shot Learning, Large Language Models.}

\maketitle
\section{Introduction}
The dynamic graph is a powerful data structure for modeling the evolving relationships among entities over time in many domains of applications, including recommender systems\cite{wu2022graph}, social networks\cite{deng2019learning}, and data center DevOps\cite{john2017service}. Anomaly edges in dynamic graphs, which refer to the unexpected or unusual relationships between entities\cite{ma2021comprehensive}, are valuable traces of almost all web applications, such as abnormal interactions between fraudsters and benign users or suspicious interactions between attacker nodes and user machines in computer networks. Due to the temporary nature of dynamics, the types of anomaly edges vary greatly, leading to the difficulty of acquiring sufficient labeled samples of new types. Therefore, detecting anomaly edges with few labeled samples plays a vital role in dynamic graph analysis and is of great importance for various applications, including network intrusions\cite{wi2022hiddencpg, balzarotti2008saner}, financial fraud detection\cite{huang2022dgraph, lu2022bright}, and \etc

\begin{figure}[t]
	\centering
	\includegraphics[width=0.48\textwidth]{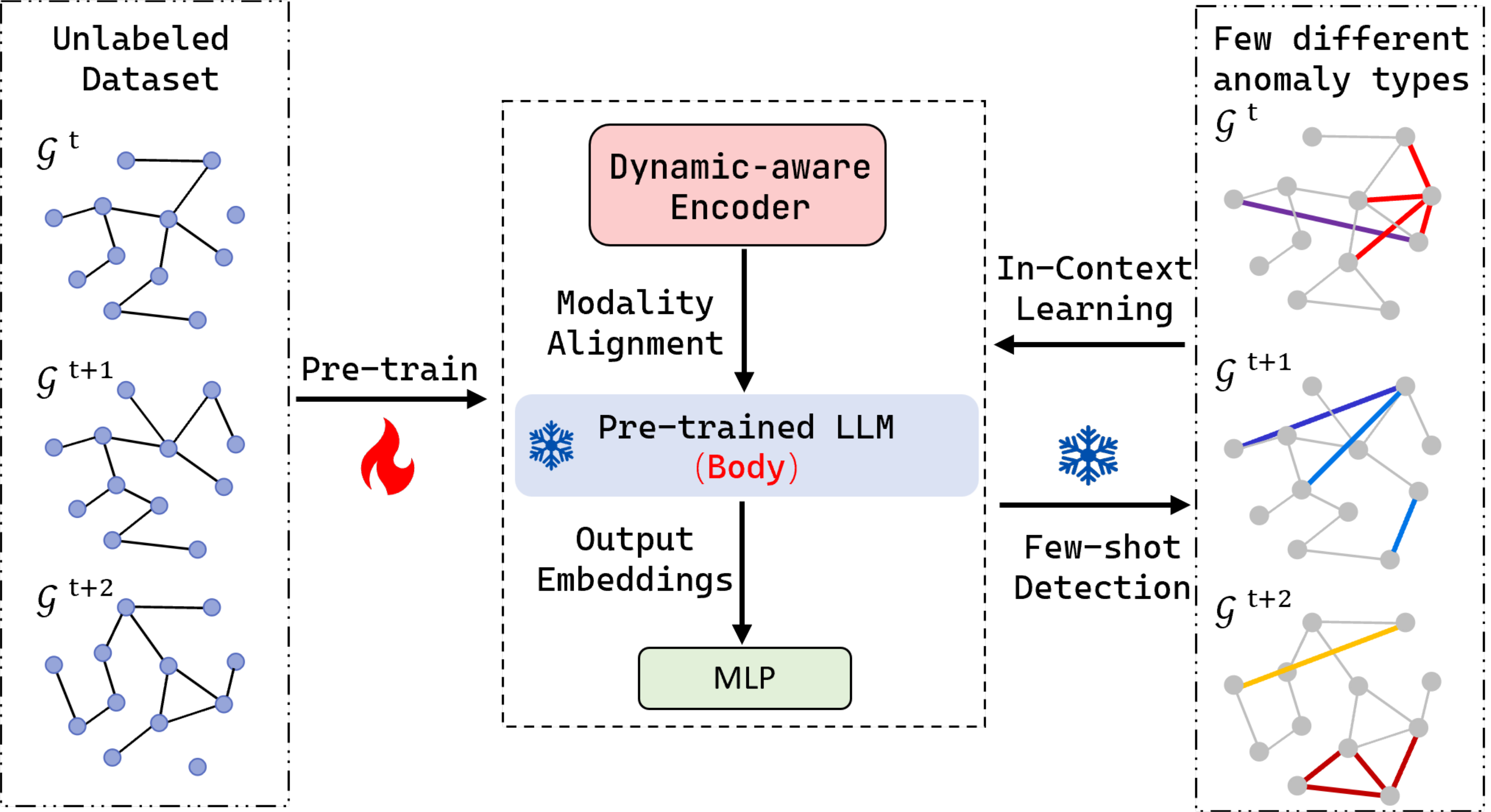}
        \vspace{-3ex}
	\caption{The motivation of \our. In the real world, edge anomaly types are diverse, evolving over time, and typically associated with limited labeled data.}
	\label{fig:motivation}
        \vspace{-3ex}
\end{figure}

Recently, various techniques have been proposed to detect anomalies in dynamic graphs. Based on the usage of labeled information, existing solutions can be categorized into three groups: supervised methods, unsupervised methods, and semi-supervised methods. Supervised methods\cite{meng2021semi, ding2021few,wang2019nodes,miz2019anomaly}utilize labeled training samples to build detectors that can identify anomalies from normal edges. Although they have demonstrated promising results, obtaining an adequate number of labeled anomaly edges for model training is challenging for dynamic graphs, which limits their scalability. Unsupervised methods\cite{duan2020aane,liu2022deep,cai2021structural,ranshous2016scalable,liu2021anomaly,zheng2019addgraph,yang2020h,ding2019deep} aim to identify anomalies in dynamic graphs without the use of label information. These approaches typically rely on statistical measures\cite{duan2020aane, liu2022deep}, graph topology\cite{cai2021structural,ranshous2016scalable}, or graph embedding techniques\cite{liu2021anomaly,zheng2019addgraph,yang2020h} to capture deviations from normal patterns. Without label information, they are mainly designed to detect randomly inserted edges as anomalies and are hard to extend for other anomaly types. Only one work, namely SAD\cite{tian2023sad}, tries to address the problem using semi-supervised learning. However, the training data used in SAD contains hundreds of labeled samples, which is also impractical in most cases. As shown in Figure \ref{fig:motivation}, with the evolution of time, the anomaly edges may change and new types of anomaly edges would emerge. For these new types, only a few (less than 10) labeled samples are available for model training. Thus, the problem we aim to solve is to identify various types of anomaly edges in the dynamic graph with few labeled samples for each type. To the best of our knowledge, there is no existing work that can be directly used for this problem. 

With the rapid progress of foundation models, large language models (LLMs) show a remarkable capability of understanding graph data\cite{tang2023graphgpt,ye2023natural} and generalizability on new tasks\cite{sun2023all}, which offers a promising path to achieve few shot anomaly edges detection for dynamic graphs. However, this task is also challenging in three aspects: (1) Representation of dynamic graph. Anomaly edges in dynamic graph are related to the changing of the graph topology. The edge representations should not only encode the information of adjacent topology but also be aware of the temporal dynamics. (2) Alignment between graph and neural language. LLMs operate on discrete tokens, whereas dynamic graphs change in continuous time. It remains an open challenge to align the semantics between dynamic graphs and word embeddings of LLMs. (3) Adaptation with few anomaly samples. To achieve few-shot detection, both LLMs and the anomaly detector should make full use of the label information of limited anomaly samples to identify different anomalies.

To solve the challenges, we proposed a novel method, namely \our, to integrate the power of LLMs and detect anomaly edges with few labeled samples. It is composed of three key modules, \ie, dynamic-aware contrastive pretraining, reprogramming-based modality alignment, and in-context learning for few-shot detection. Without using the label information, \our first employs a novel structural-temporal sampler to organize triple-wise subgraphs and pre-trains a dynamic-aware encoder of edges with contrastive loss. To align the graph encoder to LLMs, we keep the LLMs intact and reprogram the edge embeddings by text prototypes before feeding them into the frozen LLMs. Along with the reprogrammed edges, a prompt strategy is proposed to enrich the input context and direct the ability of LLMs. Both the edge embeddings and the output of LLMs are fused to identify the normal/random sampled edges. Moreover, to achieve few-shot, we employ in-context learning framework and design a prompt template that is flexible enough to encode a few labeled samples of various anomaly types. In this way, \our is able to detect different types of anomalies without modifying the model parameters. 

Compared to existing solutions, \our has the following attractive advantages: (1) \textbf{Anomaly type-agnostic.} \our conducts the dynamic graph encoding and the modality alignment in an unsupervised manner. The information of anomaly type is only used to construct the prompt of in-context learning. For detecting different anomaly types, all we need is a new prompt, \ie the model parameters are anomaly type-agnostic. (2) \textbf{Fine-tuning free.} \our directly uses the pre-trained LLMs as the backbone and keeps it intact during the reprogramming-based modality alignment. The parameters in LLMs do not require expensive fine-tuning computations. (3) \textbf{Simple to upgrade.} In \our, LLMs are only related to modality alignment parameters, and the training time for these parameters is not lengthy. If there is an alternative more powerful LLM, \our is simple to be upgraded by retraining the related parameters. The main contributions of this paper can be summarized as follows:
\begin{itemize}[leftmargin=4mm]
    \item We propose a novel method \our leveraging the advanced capabilities of LLMs for few-shot anomaly edge detection. To the best of our knowledge, this is the first work that integrates LLMs for anomaly detection of dynamic graphs.
    \item We introduce a reprogramming-based modality alignment technique, which represents the graph edge embeddings with some text prototypes, to bridge the gap between the dynamic-aware encoder and the LLMs.
    \item An in-context learning strategy is designed to integrate the information of a few labeled samples, making \our adaptable to various anomaly types with minimal computational overhead.
    \item Extensive experiments on four datasets show that \our can not only consistently outperform all baselines in few-shot detection settings but also achieve high efficiency in both alignment tuning and inference.
\end{itemize}
\vspace{-1ex}
\section{Related Work}
\begin{figure*}[ht]
	\centering
	\includegraphics[width=0.99\textwidth]{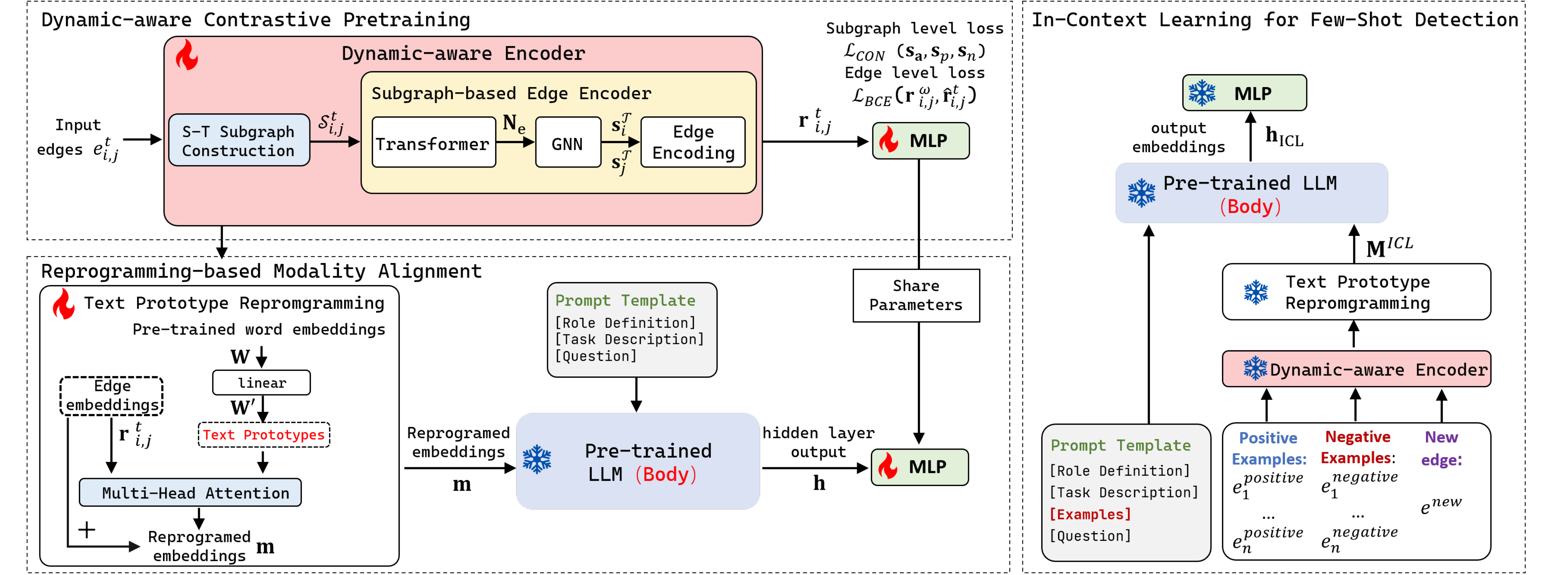}
        \vspace{-2ex}
        \caption{Overview of \our. \our comprises three modules: Dynamic-aware Contrastive Pretraining, Reprogramming-based Modality Alignment, and In-Context Learning for Few-Shot Detection.}
        \vspace{-3ex}
	\label{fig:overview}
\end{figure*}

In this section, we provide an overview of existing studies related to \our from three perspectives: (1) graph anomaly detection (2) Large Language Models (3) few-shot learning.

\textbf{Graph Anomaly Detection.} Existing graph anomaly detection methods can be broadly divided into three categories, supervised method, unsupervised method, and semi-supervised method. Most supervised methods \cite{meng2021semi,ding2021few,wang2019nodes,miz2019anomaly} rely on labeled data to train anomaly detectors, which may result in poor performance due to the limited number of samples in real-world scenarios. Unsupervised methods \cite{duan2020aane,liu2022deep,cai2021structural,ranshous2016scalable,liu2021anomaly,zheng2019addgraph,yang2020h,ding2019deep} primarily identify anomalies based on statistical measures or graph topology. These techniques mainly rely on randomly-inserted edges\cite{yu2018netwalk} during training, which differs from actual anomalies. Recently, with the advancement of semi-supervised techniques, a hybrid methods like SAD \cite{tian2023sad} have been proposed to incorporate both labeled and unlabeled data. However, these methods rely on a considerable amount of labeled samples. Nevertheless, all of these methods need the node attributes, which is not easy to obtained in dynamic graph data.

\textbf{Large Language Models.} The emergence of large language models \cite{pawelczyk2023context,fifty2023context} has ushered in a new era of few-shot learning capabilities, exemplified by their application in In-Context reasoning with minimal examples. Many LLM-based methods \cite{tang2023graphgpt,ye2023natural} are proposed to graph analysis, primarily focusing on leveraging the rich textual attributes inherent in graphs. These techniques mainly rely on modality alignment between graph representations and textual properties. However, this reliance significantly limits their applicability in scenarios where textual attributes are absent. While some efforts \cite{jin2023time} have been made to enhance LLMs' understanding of non-textual data like time-series, through reprogramming techniques, the application of these methodologies to graph data, especially dynamic graphs, remains largely unexplored.

\textbf{Few-shot Learning in Dynamic Graphs.} The challenge of limited labeled data is pervasive in real-world applications. Many studies have explored for few-shot learning, using techniques like meta-learning or contrastive learning \cite{wang2019semi, wei2022contrastive, zhao2023doubleadapt, zhu2023wingnn, xu2023metagad, liu2021relative, ding2020graph}. However, these methods are generally tailored to static graphs or specific tasks \cite{guo2021few, li2020few}, leaving a gap in anomaly edge detection for dynamic graphs. Our study addresses this gap by leveraging the potential of LLMs in a few-shot learning context for anomaly detection in dynamic graphs.

\vspace{-1em}
\section{Preliminary}
\subsection{Problem Definition}
Let $\mathcal{G} = [\mathcal{G}^{1}, ..., \mathcal{G}^{t}, ..., \mathcal{G}^{T}]$ denote a sequence of graph snapshots spanning timestamps $1$ to $\mathcal{T}$, where each snapshot $\mathcal{G}^{t} = (\mathcal{V}^{t}, \mathcal{E}^{t})$ represents the state of the graph at time $t$ with $\mathcal{V}^{t}$ being the set of nodes and $\mathcal{E}^{t}$ the set of edges. An edge $e_{i, j}^{t} = (v_{i}^{t}, v_{j}^{t}) \in \mathcal{E}^{t}$ signifies an interaction between nodes $v_{i}^{t}$ and $v_{j}^{t}$ at time $t$. The structure of each snapshot is encoded in a binary adjacency matrix $\mathbf{A}^{t} \in \mathbb{R}^{n \times n}$, where $\mathbf{A}^{t}_{i,j} = 1$ if there is an edge between $v_i$ and $v_j$ at timestamp $t$, and $\mathbf{A}^{t}_{i,j} = 0$ otherwise.

Considering the high cost of acquiring large-scale labeled anomaly samples in real-world scenarios, we focus on detecting anomaly edges leveraging only a minimal amount of labeled data. Note that we assume the nodes in $\mathcal{G}$ are relatively stable. Given a specified anomaly type $\mathcal{T}$ and related set of few anomaly edges $\mathcal{E}_{\mathcal{T}}= \{\mathcal{T}_1, \cdots, \mathcal{T}_{a}\}$, where $a$ is the number of anomaly edges, our objective is to detect whether edge $e_{i, j}^t$ in $\mathcal{G}^{t}$ is an anomaly edge of type $\mathcal{T}$ or not.

\subsection{Overview of \our}
As shown in Figure \ref{fig:overview}, \our is a LLM enhanced few-shot anomaly detection framework. It consists of three key modules: dynamic-aware encoder, modality alignment and in-context learning for detection. 

\begin{itemize}[leftmargin=4mm]
    \item Given an edge $e_{i,j}^t$, the dynamic-aware encoder captures the related temporal and structure information from the dynamic graph, and encodes it into the edge representation. We construct a series of structural-temporal subgraphs $\mathcal{S}_{i,j}^t$ of edge $e_{i,j}^t$. Based on these subgraphs, \our generates the edge embedding $\mathbf{r}$ by fusing all the related subgraphs in $\mathcal{S}_{i,j}^t$.
    \item Taking $\mathbf{r}$ as the input, we first select some dynamic graph-related words and cluster them into $\textbf{V}'$ prototypes. \our adopt self-attention to reprogram the edge embedding $\mathbf{r}$ with the textual prototype and obtain $\mathbf{h}$. Both existing edges and randomly selected edges are employed to construct pseudo labels for alignment fine-tuning. 
    \item For few-shot detection, we utilize in-context learning to encode the label information from a few anomaly samples. A prompt template consisting of role definition, task description, examples and questions is designed to 
    embed the edge representations $\mathbf{h}$ and detect various types of anomalies without any update of model parameters.

\end{itemize}

\vspace{-1em}
\section{Methodology}

As shown in Figure \ref{fig:overview}, \our consists of three key modules, \ie, dynamic-aware contrastive pretraining, reprogramming-based modality alignment, and in-context learning for few-shot detection. Next, we specify the details of each module respectively.

\subsection{Dynamic-aware Contrastive Pretraining}
Dynamic graphs are changing over time, leading to the difficulty in representing the structure and temporal information of the edges. Existing solutions either focus on the structure information by averaging the context of adjacent nodes\cite{zheng2019addgraph}\cite{cai2021structural} or directly use sequential models to capture the temporal dynamics\cite{liu2021anomaly}\cite{yu2018netwalk}, which are not sufficient for the anomaly detection. In this section, we propose the dynamic-aware contrastive pretraining to systematically model both aspects and represent the edges with their adjacent subgraphs. The whole module consists of two subparts, \ie dynamic-aware encoder and contrastive learning-based optimization.

\subsubsection{Dynamic-aware Encoder} Given an edge $e_{i,j}^t$, we first construct structrual-temporal subgraphs $\mathcal{S}_{i,j}^t$, then fed it into the subgarph-based edge encoder to obtain the edge representation $\mathbf{r}_{i,j}^t$.

\textbf{Structural-Temporal Subgraph Construction. }\label{sec:sampler}
For an edge $e_{i,j}^t$, we design to construct structural-temporal subgraphs for both source and target nodes. 
Given an edge $e_{i, j}^{t} = (v_{i}^{t}, v_{j}^{t}) \in \mathcal{E}^{t}$, we first construct a diffusion matrix\cite{liu2021anomaly} $\textbf{D}^t\in \mathbb{R}^{N \times N}$ of $\mathcal{E}^{t}$ to select the structure context, where $N$ represents the number of nodes in $\mathcal{E}^{t}$.

Each row $d_{i}^t$ of $\textbf{D}^t $ indicates the connectivity strength of the $i-th$ node with all other nodes in the graph $\mathcal{G}^t$. For $e_{i, j}^{t} = (v_{i}^{t}, v_{j}^{t})$, we utilize $d_{i}^t$ and $d_{j}^t$ to select the most significant top-$K$ adjacent nodes of $\mathcal{V}^{t}$ to form $\mathcal{V}_{i}^t$ and $\mathcal{V}_{j}^t$ as the subgraph nodes of the source node $v_i^t$ and target node $v_j^t$ respectively. Then, we link the nodes in $\mathcal{V}_{i}^t$ to its related node $v_i^t$ to generate $\mathcal{E}_{i}^t$ and obtain the subgraphs $\mathcal{g}_{i}^t = \{\mathcal{V}_{i}^t , \mathcal{E}_{i}^t\}$. Similar operations are conducted for the target node $v_j^t$ to obtain $\mathcal{g}_j^t = \{\mathcal{V}_{j}^t, \mathcal{E}_{j}^t\}$. In this way, both the source and the target in $e_{i,j}^t$ can be represented by the relevant surrounding subgraphs $\mathcal{g}_{i,j}^t = [\mathcal{g}_i^t, \mathcal{g}_j^t$]. 

To obtain the temporal context of $e_{i,j}^t$, \our utilizes a sliding window $\Gamma$ to filter a sequence of graph slices $\mathcal{G}_{t}^{\Gamma} = \{\mathcal{G}_{t-\Gamma+1}, $ $\ldots, \mathcal{G}_{t}\}$. For each graph slice, we use the described method to construct subgraphs. Therefore, a sequence of subgraph for $e_{i,j}^t$ can be constructed as follows:
\begin{equation*}
    \mathcal{S}_{i,j}^t = \{\mathcal{g}_{i,j}^\tau\} \quad \text{for } \tau = t-\Gamma+1, \ldots, t
\end{equation*} 
$\mathcal{S}_{i,j}^t$ contains not only the structure but also the temporal context of $e_{i,j}^t$. The representation of $\mathcal{S}_{i,j}^t$ can be used to detect the anomaly in $\mathcal{G}$.

\textbf{Subgraph-based Edge Encoder. }
Given the subgraph sequence $\mathcal{S}_{i,j}^t$ of edge $e_{i,j}^t$, we feed them into the subgraph-based edge encoder which synergizes the Transformer and Graph Neural Network (GNN) models to obtain edge representation $\mathbf{r}_{i,j}^t\in \mathbb{R}^{d_{m}}$, where $d_{m}$ represents the embedding dimension. 
Following the same setting as Taddy\cite{liu2021anomaly}, we assume the nodes in $\mathcal{G}$ are stable and conduct the following four steps on the input  $\mathcal{S}_{i,j}^t$: 
\begin{itemize}[leftmargin=4mm]
    \item \textbf{Node Encoding.} For each node $v_l^\tau$ in every $\mathcal{g}_i^\tau$ within $\mathcal{S}_{i,j}^t$, we construct the node encoding using three aspects, \ie, $\mathbf{z}_l=\mathbf{z}_{\mathrm{diff}}(v_l^\tau)+\mathbf{z}_{\mathrm{dist}}(v_l^\tau)+\mathbf{z}_{\mathrm{temp}}(v_l^\tau) \in \mathbb{R}^{d_{enc}}$. Here, $\mathbf{z}_{\mathrm{diff}}(v_{l}^\tau)$ represents the diffusion-based spatial encoding capturing the global structural role of node $v_l^\tau$, $\mathbf{z}_{\mathrm{dist}}(v_l^\tau)$ denotes the distance-based spatial encoding, reflecting the local structural context; and $\mathbf{z}_{\mathrm{temp}}(v_l^\tau)$ provides the relative temporal information of node $v_l^\tau$ which is the same for all nodes at the time slice $\tau$. 
    \item \textbf{Temporal Encoding.} We model the temporal information of nodes in $\mathcal{S}_{i,j}^{t}$ by reorganizing the node encoding into an encoding sequence $\mathbf{Z}_e = [[\mathbf{z}_l]_{v_l \in g_{i,j}^{\tau}}]_{g_{i,j}^{\tau} \in \mathcal{S}_{i,j}^{t}}$, with the dimension of $\mathbf{Z}_e $ being $\mathbb{R}^{(2(K+1)\cdot \Gamma) \times d_{enc}}$.  We feed $\mathbf{Z}_e$ into a vanilla Transformer block to obtain the node embeddings $\textbf{N}_e = \rm Transformer(\mathbf{Z}_e)$. The dimension of node embedding, $d{enc}$, is specified here.
    \item \textbf{Subgraph Encoding.} Additionally, we employ GNN to generate the graph representations of all related subgraphs in $\mathcal{S}_{i,j}^{t}$. For each subgraph $g_i^\tau$, we extract the related node embeddings $\textbf{N}_i^t\in \mathbb{R}^{(K+1) \times d_{enc}}$ from $\textbf{N}_e$ and utilize GNN to obtain the embedding of node $v_l^\tau$ as the subgraph embedding $\textbf{s}_i^\tau$. To fuse the information on different timesteps, we stack the $\Gamma$ embeddings of $v_l^\tau$ to generate $\textbf{s}_i^\tau \in \mathbb{R}^{(K+1) \times d_{enc}}$
    \item \textbf{Edge Encoding.} To obtain the representation of $e_{i,j}^t$, we first conduct average pooling on the related subgraph embeddings $\mathbf{s}_i^\tau$ and $\mathbf{s}_j^\tau$. Subsequently, we concatenate the resulting vectors and project the concatenated vector into the LLM's hidden dimension $d$ using a fully connected layer. The final representation of $\mathbf{r}_{i,j}^t$ is thus given by
    \begin{equation*}
        \mathbf{r}_{i,j}^t = \text{fc}(\text{concat}(\text{AvgPool}(\mathbf{s}_i^\tau), \text{AvgPool}(\mathbf{s}_j^\tau)))  \quad \text{for } \tau = t-\Gamma+1, \ldots, t
    \end{equation*} 

\end{itemize}
By incorporating this step, \our can systematically model the structural and temporal dynamics. More details of the subgraph-based edge encoder can be found in the Appendix ~ \ref{appen:detail}.

\subsubsection{Contrastive Learning-based Optimization}
\label{sec:constructive-Learning}
\our employs contrastive learning to optimize the parameters in the dynamic-aware encoder. To obtain negative samples and achieve anomaly detection, we follow two principles in sampling: (1) edges with different subgraphs of related nodes should not have similar embeddings; (2) the embeddings between existing edges and randomly sampled edges should be distinguishable. 

\begin{figure}[t]
	\centering
	\includegraphics[width=0.4\textwidth]{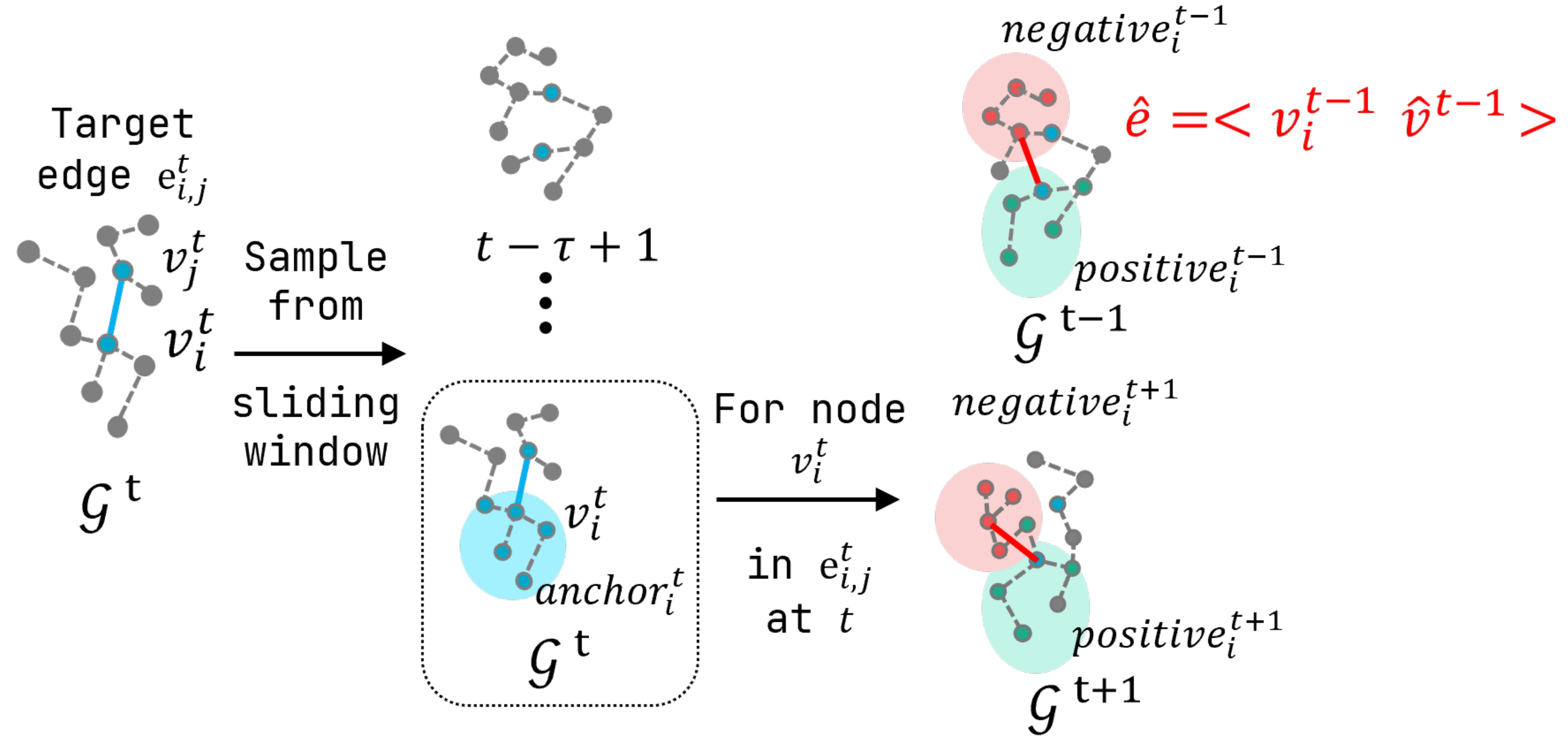}
        \vspace{-2ex}
	\caption{Sample process of contrastive training triplet}
        \vspace{-3ex}
	\label{fig:sampling}
\end{figure}

For edge $e_{i,j}^t$, we check its adjacent graphs, $\mathcal{G}^{t-1}$ and $\mathcal{G}^{t+1}$. The sampling should include two levels, \ie edge level and subgraph level. As shown in Figure \ref{fig:sampling}, we randomly sample a node $\hat{v}^{\omega}, \text{ where } \omega = t \pm 1$ not directly connected to $v_i^{\omega}$ and generate the edge embedding $\hat{\mathbf{r}}_{i,j}^t$ for the edge $<v_i^{\omega}, \hat{v}^{\omega}>$. At the edge level, we employ a Multilayer Perceptron (MLP) layer as the anomaly detector to identify whether the input edge is randomly sampled. Here, we feed the embeddings of $\mathbf{r}_{i,j}^{\omega}$ and $\hat{\mathbf{r}}_{i,j}^t$ into the detector and employ binary cross-entropy loss to make them distinguishable:
 $$\mathcal{L}_{BCE} = -\log(1- \rm{MLP}(\mathbf{r}_{i,j}^t)) + \log(\rm{MLP}(\mathbf{r}_{i,j}^t))$$

At the subgraph level, we consider the subgraph of $\hat{v}^{\omega}$ as the negative sample of subgraph of $v_i^t$ and utilize the subgraph of $v_i^{\omega}$ in different timestamps as the positive sample to construct a triplet. As shown in the right part of Figure \ref{fig:sampling}, we sample negative subgraph and contrastive training triplet for node $v_i^{\omega}$. Since the edge embeddings are concatenations of subgraph embeddings, \our employ contrastive loss to enlarge the dissimilarity between subgraph embeddings, and the pretraining loss is the combination of both edge level loss and subgraph level loss. 
\begin{align}\label{eq:pretrain-loss}
    \mathcal{L}_{con} &= -\log \frac{\exp(\cos(\textbf{s}_a, \textbf{s}_p) / \delta)}{\exp(\cos(\textbf{s}_a, \textbf{s}_p) / \delta) + \exp(\cos(\textbf{s}_a, \textbf{s}_n) / \delta)}\\
    \mathcal{L} &= \mathcal{L}_{BCE}+\mathcal{L}_{con}
\end{align}
where $ \textbf{s}_a, \textbf{s}_p , \textbf{s}_n \in \mathbb{R}^{d_{emb}}$ represent the subgraph embeddings for the anchor, positive, and negative samples in the triplet. $\cos()$ denotes the cosine similarity between two sample embeddings, and $\delta$ is a temperature parameter that controls the scale of the similarity scores.
\subsection{Reprogramming-based Modality Alignment}
\label{sec:text-prototype-alignment-and-tuning}

For few-shot detection, the representations of edges should be general enough to be adapted to various anomaly types with few labeled samples. \our employs LLMs as the backbone to enhance the generalization ability of edge embeddings output by the dynamic-aware encoder. This is rather challenging because of the modality difference between dynamic graphs and neural languages. Thus, we propose reprogramming-based modality alignment techniques to bridge the gap. For simplicity, we omit the subscript and note the edge embedding with $\textbf{r}$. Taking the $\textbf{r}$ as input, \our first reprograms it with the prototype of the word embeddings and feeds the reprogramed vector into LLMs to generate $\textbf{h} \in \mathbb{R}^{d}$. Both $\textbf{r}$ and $\textbf{h}$ are fused as the final edge embedding to input to the LLM for anomaly detection.

\subsubsection{Text Prototype Reprogramming}
\label{sec:text-prototype-alignment}
Although LLMs are trained with neural languages, the learned parameters contain the knowledge of almost all domains and can be viewed as a world model\cite{hao2023reasoning}. To leverage the capability of LLM for dynamic graph analysis, we first select a subset of word embeddings and cluster them as text prototypes for reprogramming edge embeddings.

Specifically, given the pre-trained word embeddings of LLMs, we refine a subset of words \( \textbf{W}\in\mathbb{R}^{V \times d} \) related to dynamic graphs to generate text prototypes. In practice, we prompt the LLM with a question, \ie \texttt{Please generate a list of words related to dynamic graphs to align dynamic graph data with natural language vocabulary}. The full version of this question can be found in the Appendix ~\ref{appen:prompt}. The output words in different rounds are combined to obtain $V$ related words. Based on these words, we construct the text prototype with liner transformation:
$$
\textbf{W}' = \textbf{M} \cdot \textbf{W}
$$ 
where $\textbf{M}\in \mathbb{R}^{V' \times V}$ and $V'$ is the number of prototypes. Given an edge embedding $\textbf{r}$, \our utilize multi-head cross-attention to conduct reprogramming. We use $\textbf{r}$ as the query vector and employ $\textbf{W}'$ as the key and value matrices. For each attention head \( c \) in \( \{1, \ldots, C\} \), we compute the related query, key and value matrices, \ie, $\textbf{Q}_c$, $\textbf{K}_c$ $\textbf{V}_c$. The attention operation for each head is formalized as:
\begin{align*}
\textbf{z}_c = \text{A\scalebox{0.8}{TTENTION}}(\textbf{Q}_c, \textbf{K}_c, \textbf{V}_c)
\end{align*}

The outputs from all heads are aggregated to obtain $\textbf{z} \in \mathbb{R}^{d}$. We then add $\textbf{z}$ to the edge embedding $\textbf{r}$ to obtain the reprogramed representation $\textbf{m}\in \mathbb{R}^{d}$ of the given edge $e_{i,j}^t$. 
\subsubsection{Pseudo Label for Anomaly Fine-tuning}
\label{sec:model-tuning}

In \our, the backbone LLM takes the reprogrammed input $\textbf{m}$ as input to generate the final representation vector for anomaly detection. Since the parameters of LLMs are intact, the representation of LLM may not contain the information on edge anomalies and may not suit for few-shot detection. Therefore, we utilize the randomly sampled edges (detailed in Section~\ref{sec:constructive-Learning}) as pseudo anomaly labels to fine-tune the parameters of the dynamic-aware encoder and anomaly detector.

\begin{figure}[t]
	\centering
	\includegraphics[width=0.49\textwidth]{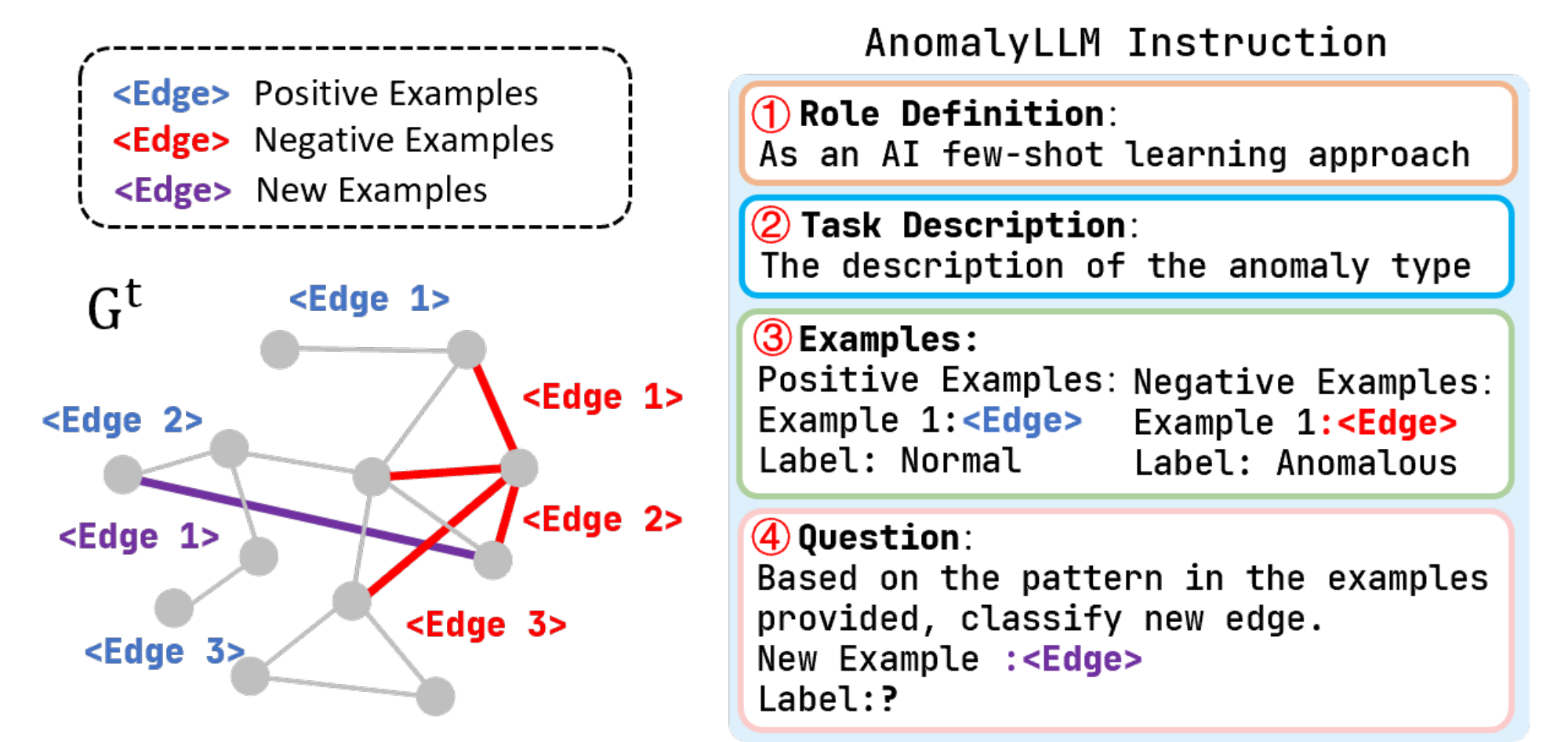}
        \vspace{-3ex}
	\caption{The prompt of In-Context Learning} 
	\label{fig:prompt}
 \vspace{-3ex}
\end{figure}

As shown in Figure \ref{fig:prompt}, we design a template of prompt for both alignment fine-tuning and in-context learning detection. The template consists of four aspects: role definition, task description, examples and questions, where <Edge> is a mask token for the input edge embedding. We detail the prompt in Section~\ref{sec:template}. 
The instruction is fed into the LLM and the hidden state of the <Edge> token is selected as the final representation vector of edge $e$. For conciseness, we use $v$ to represent $v_{i,j}^t$. This procedure can be formalized as follows:
\begin{equation*}
\textbf{H} = \textbf{LLM}([\mathbf{u}, \textbf{m}])
\end{equation*}
where $\mathbf{u}\in \mathbb{R}^{L \times d}$ is the related embeddings of instruction templates and  $\textbf{H}\in \mathbb{R}^{(L+1) \times d}$ is the last hidden layer output of the LLM. We utilize the last position of $\textbf{H}$, \ie $\textbf{h}$ for detection. Note that our backbone LLM employs causal attention to compute $\textbf{h}$. Thus, for different edges, the front parts of $\textbf{h}$ are the same. We can use this character to further reduce the computation workload in the pre-training procedure.
 
As described in Section~\ref*{sec:constructive-Learning}, an MLP layer is employed to detect the randomly selected anomalies and output an anomaly score for input edge embedding. In this module, we reuse the MLP detector and replace the input edge embedding $\textbf{r}$ with the reprogramed edge embedding $\textbf{r}$. The anomaly score for an edge $e$ is computed with $f(e) = \rm MLP(\textbf{h})$. We also used the randomly selected edges as negative samples and the existing edges as positive samples to construct pseudo labels. A binary cross-entropy (BCE) loss of pseudo labels is employed to optimize the parameters of the dynamic-aware encoder and the detector. $\mathcal{L}_{BCE} = -\log(1- f(e)) + \log(f(e))$
Note that the MLP detector is optimized in both pre-training and alignment fine-tuning. In few-shot anomaly detection, the MLP detector cooperates with the in-context learning strategy to detect various types of anomalies. During the whole procedure, the parameters of LLM are intact.

\subsection{In-Context Learning for Few-Shot Detection}
\label{sec:in-context-learning-few-shot}
Given a set of anomaly edges $\mathcal{E}_{\mathcal{T}}= \{\mathcal{T}_1, \cdots, \mathcal{T}_{a}\}$ of anomaly type $\mathcal{T}$, \our aim to detect whether the new edge $e$ is an anomaly edge of $\mathcal{T}$ or not. Considering that the pretraining procedure of \our has no information about the anomaly type, we need to make full use of the labeled information of $\mathcal{E}_{\mathcal{T}}$. In this paper, we proposed to use in-context learning that encodes edges in $\mathcal{E}_{\mathcal{T}}$. Next, we introduce the construction of the prompt template and few-shot anomaly detection respectively.
\subsubsection{Prompt Template Construction}\label{sec:template}
The ability of LLMs on downstream tasks can be unleashed by in-context learning which learns from the context provided by a prompt without any additional external data or explicit retraining. Thus, how to construct the prompt template is a critical problem. In \our, we argue that the prompt should contain the information of four aspects: role definition, task description, examples and question. 

As shown in Figure~\ref{fig:prompt}, the prompt first defines the role of LLM as a few-shot anomaly detector followed by the description of anomaly type $\mathcal{T}$. For the example part, we select the same number of edges $\mathcal{E}'/ \mathcal{E}_{\mathcal{T}}^{'}$ from $\mathcal{E} / \mathcal{E}_{\mathcal{T}}$ as the normal and anomaly samples and generate the embedding of edges in $\mathcal{E}' \cup \mathcal{E}^{'}_{\mathcal{T}}$ with dynamic-aware encoder denoted by $\textbf{M}^{ICL}$. These edges are then processed through the reprogramming module for modality alignment and to build the prompt examples. 
\begin{align*}
	\textbf{M}^{ICL} &= \{\textbf{m}^{pos}_1, \ldots, \textbf{m}^{pos}_n, \textbf{m}^{neg}_1, \ldots, \textbf{m}^{neg}_P, \textbf{m}^{new}\} \\
	\textbf{h}_{ICL} &= \mathbf{LLM}([\textbf{u}_{ICL}, \textbf{M}^{ICL}])[:-1];\
	f(e^{new})=\rm MLP(\textbf{h}_{ICL})
\end{align*}
where $\textbf{m}^{pos}_u,\textbf{m}^{neg}_u\in \mathbb{R}^{d_m}$ are the reprogrammed embeddings of the $u$-th positive and negative edge examples, respectively, and $\textbf{m}^{new}\in \mathbb{R}^{d_m}$ is the reprogrammed embedding of the edge under investigation. In the prompt template, we employ mask token <Edge> to represent the location of edge embeddings and each example has a related label tag to make use of the given few labeled data. Given a new edge $e^{new}$ needed to be detected, we conduct the same operations of examples to obtain the edge embedding.

\subsubsection{Few-shot anomaly detection}
Using \our, we can conduct few-shot anomaly edge detection for various anomaly types without any update of parameters. For a specific anomaly type $\mathcal{T}$, the ICL template can be constructed in advance. Assuming $e^{new}$ is a new edge to be detected, \our utilize the dynamic-aware encoder to obtain an intermediate vector and reprogram it with text prototypes. By embedding the reprogrammed vector into the ICL template, we obtain the input of LLM to generate the edge embedding $\mathbf{h}_{ICL}$. Then, the edge embedding is fed into the pre-trained anomaly decoder, \ie the MLP layer, to calculate the probability of $e^{new}$ to be an anomaly of $\mathcal{T}$:
$$
f(e^{new}) = \rm MLP(\mathbf{h}_{ICL})
$$
For different anomaly types, we can build multiple ICL templates by using a few labeled samples for each type. The reprogrammed vector of $e^{new}$ is embedded in these templates to generate the edge embedding and the anomaly probability of various anomaly types. Due to the causal attention mechanism of our backbone LLM, both the embedding of a few labeled edges and the intermediate embedding of ICL templates can be precomputed in advance. Once the reprogramed vector is generated, \our conducts constant operations to obtain the anomaly probability, leading to high efficiency. Next, we further analyze the complexity of \our to illustrate this character.

\subsection{Complexity Analysis of \our}
\label{sec:complexity-analysis}
Due to the limitations of space, we only analyze the inference complexity here. The complexity of model training is detailed in the Appendix~\ref{appen:prompt}. Given the well-optimized model, \our involve four parts to detect an edge $e^t_{i,j}$, \ie, subgraph construction, dynamic-aware embedding computation, reprogramming and ICL inference of LLM. 
\begin{itemize}[leftmargin=4mm]
	\item For subgraph construction, \our select $K$ related nodes for nodes $v_i$ and $v_j$. Cause the diffusion matrix of $\mathcal{G}$ at all timestamps can be precomputed, the complexity of this part is $O(\Gamma \times K)$ where $\Gamma$ is the temporal window size. 
	\item For dynamic-aware embedding, \our takes the nodes in the subgraphs as input and compute the $\mathbf{z}_{\mathrm{diff}}(v_{i})$, $\mathbf{z}_{\mathrm{dist}}(v_{i})$ and $\mathbf{z}_{\mathrm{temp}}(v_{i})$ for each node $v_{i}$ as the node features. The complexity of this part is $O(3d)$. Then, the sequence of node features is fed to the Transformer block to obtain node embeddings, with the complexity of $O((2(K+1)\Gamma)^2d + 2(K+1)\Gamma d^2)$. A GNN layer and average pooling layer of subgraphs is conducted on these embeddings to generate the dynamic-aware embedding $\textbf{r}_{i,j}$, and the complexity is $O((K+1)^2\Gamma d)$. Therefore, the complexity of this part is $O((K+1)^2\Gamma^2d + (K+1)\Gamma d^2)$.
	\item \our utilizes self attention to reprogram $\textbf{r}_{i,j}$ and generate $\textbf{m}$. The complexity is $O(V'd + V'd^2) = O(V'd^2)$.
	\item Due to the causal attention of LLM, the hidden states of the ICL template are the same except for the last <Edge> embedding $\textbf{h}$. Thus, for the inference of LLM, \our precomputes and stores the intermediate hidden state of ICL template, and directly conducts $O(Y)$ feed-forward operations to obtain $\mathbf{h}$, where $Y$ is the number of Transformer layers in LLM.
\end{itemize}
According to the analysis, the complexity of detecting $e^t_{i,j}$ is the summarization of the four parts. Note that $\Gamma$, $K$, $d$, $V'$ and $L$ are constant for \our, the inference complexity to detect $e^t_{i,j}$ is also a constant.
\section{Experiments}

\begin{table*}[t]
  \caption{ Performance comparison results of few-shot anomaly detection on multiple anomaly types.}
  \vspace{-3ex}
\begin{tabular}{c|c|ccc|ccc|ccc}         
 \hline
  \multirow{2}{*}{Dataset} & \multirow{2}{*}{Model} & \multicolumn{3}{c|}{1-shot} & \multicolumn{3}{c|}{5-shot} & \multicolumn{3}{c}{10-shot}    \\ 
 \multirow{2}{*}{} &\multirow{2}{*}{} &{CDA} & {LPL} & {HHL}& {CDA} & {LPL} & {HHL}& {CDA} & {LPL} & {HHL}  \\ 
 \hline         
  \multirow{6}{*}{} & {StrGNN}& {0.5891} & {0.5756} & {0.5974}& {0.6018} & {0.6041} & {0.6122}&{0.6222} & {0.6329} & {0.6402} \\ 
   \multirow{6}{*}{BlogCataLog} & {AddGraph}&{0.5994} & {0.6023} & {0.5988}& {0.6097} & {0.6033} & {0.6104} &{0.6216} & {0.6238} & {0.6172}  \\ 
    \multirow{6}{*}{} & {Deep Walk}&{0.6102} & {0.6073} & {0.6202}& {0.6113} & {0.6122} & {0.6196} & {0.6155} & {0.6176} & {0.6154} \\ 
    \multirow{6}{*}{} & {TGN}&{0.6732} & {0.6699} & {0.6919}& {0.7112} & {0.7023} & {0.7118} & {0.7263} & {0.7387} & {0.7311} \\ 
    \multirow{6}{*}{} & {GDN}&{0.6733} & {0.6795} & {0.6609}& {0.6997} & {0.7051} & {0.7121} &{0.7321} & {0.7311} & {0.7319}  \\ 
    \multirow{6}{*}{} & {SAD}&{0.6841} & {0.6792} & {0.6411}& {0.7002} & {0.7018} & {0.6988} & {0.7342} & {0.7216} & {0.7265} \\ 
    \multirow{6}{*}{} & {TADDY}&{0.6892} & {0.6983} & {0.6891}& {0.7148} & {0.7186} & {0.7177} & {0.7258} & {0.7326} & {0.7334} \\ 
  \multirow{6}{*}{} & \textbf{\our}&\textbf{0.8288} & \textbf{0.8334 }&\textbf {0.8255}& \textbf{0.8331} & \textbf{0.8319} & \textbf{0.8407} &\textbf{0.8402} & \textbf{0.8456} & \textbf{0.8447}\\ 
  
  \hline
  \multirow{6}{*}{UCI} & {StrGNN}&{0.6143} & {0.5956} & {0.5722}& {0.6113} & {0.7132} & {0.6512}&{0.6442} & {0.6724} & {0.6249}   \\ 
   \multirow{6}{*}{Message} & {AddGraph}&{0.5842} & {0.5466} & {0.5647}& {0.6018} & {0.6667} & {0.6321} &{0.4642} & {0.5728} & {0.7001}  \\ 
   \multirow{6}{*}{} & {Deep Walk}&{0.6198} & {0.6187} & {0.6142}& {0.6256} & {0.6263} & {0.6176} & {0.6255} & {0.6209} & {0.6197} \\ 
  \multirow{6}{*}{} & {TGN}&{0.6521} & {0.6535} & {0.6643}& {0.7098} & {0.7193} & {0.7155} & {0.7335} & {0.7365} & {0.7324} \\ 
  \multirow{6}{*}{} & {GDN}&{0.6577} & {0.6818} & {0.6611}& {0.7201} & {0.7289} & {0.7255} &{0.7493} & {0.7511} & {0.7546}  \\  
  \multirow{6}{*}{} & {SAD}&{0.6703} & {0.6587} & {0.6693}& {0.7102} & {0.7146} & {0.7194} & {0.7416} & {0.7453} & {0.7406} \\ 
    \multirow{6}{*}{} & {TADDY}&{0.6992} & {0.7078} & {0.6132}& {0.7204} & {0.7237} & {0.7218} & {0.7255} & {0.7278} & {0.7243} \\ 
  \multirow{6}{*}{} & \textbf{\our}&\textbf{0.8414} & \textbf{0.8358} & \textbf{0.8368}& \textbf{0.8446} & \textbf{0.8459} & \textbf{0.8424} & \textbf{0.8488} & \textbf{0.8546} & \textbf{0.8442} \\ 

  \hline
 \end{tabular}
 \vspace{-3ex}
   \label{tab:few-shot}
 \end{table*}

In this section, we conducted extensive experiments on \our to answer the following research questions:
\begin{itemize}[leftmargin=4mm]
    \item \textbf{Q1:} What is the performance of \our in detecting different types of anomaly with few labeled anomaly edges for each type? 

    \item \textbf{Q2:} How efficient of \our in model alignment and anomaly detection? 
    \item \textbf{Q3:} What are the influences of the proposed modules and different backbone LLMs? 

    \item \textbf{Q4:} What is the performance of \our on real-world anomaly edge detection task? 

\end{itemize}
Besides, we also studied the sensitivity of key parameters and the performance comparison on unsupervised anomaly edge detection. Due to the space limit, the results of these experiments are illustrated in Appendix~\ref{appen:parameters} and ~\ref{append:unsupervised}. All the code and data are available at \underline{https://github.com/AnomalyLLM/AnomalyLLM}.

\subsection{Experimental Settings}
We briefly introduce the experimental settings below. The detailed experimental settings can be found in the Appendix~\ref{appen:setting}.

\subsubsection{Data Descriptions.}
We use four public dynamic graph datasets to evaluate the performance of \our. The main experiments are conducted on two widely-used benchmark datasets, \ie, UCI Messages \cite{opsahl2009clustering} and Blogcatalog\cite{tang2009relational}. To evaluate the performance of \our on real-world anomaly detection task and test the capability of \our, we also employ two datasets with real anomaly, \ie T-Finance\cite{tang2022rethinking} and T-Social\cite{tang2022rethinking}, which have over $21$ million and $73$ million edges respectively.

\subsubsection{Experimental Protocol.} In this paper, we utilize both synthetic anomaly and real anomaly to evaluate the performance of \our. Existing dynamic graphs either have no labeled anomaly edges or only have one anomaly type. To verify the ability of \our on various anomaly types, we follow the experiments of \cite{liu2023rgse} and generate three kinds of systematic anomaly types, \ie, Contextual Dissimilarity Anomaly(CDA), Long-Path Links (LPL) and Hub-Hub Links(HHL) for UCI Messages and Blogcatalog datasets. The details of anomaly generation are described in Appendix ~\ref{appen:setting}. For dynamic graphs having labeled anomaly, such as T-Finance and T-Social, we directly used the real anomaly label to conduct the experiments. In our experiments, we employ all nodes and edges to pretrain the dynamic-aware encoders and align them to the backbone LLMs. For anomaly detection, only a few labeled edges are available. We build $1$-shot, $5$-shot and $10$-shot labeled edges for each anomaly type to obtain the AUC results on other edges.

\subsubsection{Baselines.}\label{sec:baseline} We compare \our with seven baselines which can be categorized into three groups, \ie, general graph representation method, unsupervised anomaly detection methods, and semi-supervised anomaly detection methods. For the first group, we select \textbf{DeepWalk}\cite{perozzi2014deepwalk} to generate the representations of edges. For unsupervised method, we employ the recent three works, \ie  \textbf{StrGNN}\cite{cai2021structural}, \textbf{AddGraph}\cite{zheng2019addgraph}, and \textbf{TADDY}\cite{liu2021anomaly}, as our baselines. For semi-supervised methods, we use \textbf{GDN}\cite{ding2021few}, \textbf{TGN}\cite{xu2020inductive}and \textbf{SAD}\cite{tian2023sad} for performance comparison. The details of how to use these methods on our tasks are specified in the Appendix \ref{appen:baseline}.

\subsubsection{Hyperparameters setting.} For subgraph construction, we set the number $k$ to be $14$ and $\Gamma$ is $4$. For edge encoder, the embedding dimension $d$ is $512$. \our employs $3$-layers stack of Transformer. We train UCI Messages, BlogCatalog, T-Finance and T-Social datasets with $150$ epochs.During the modality alignment, we fine-tune the encoder and anomaly detector for $20$ epoch. All the experiments are conducted on the $2\times$Nvidia 3090Ti. 

\subsection{Performance Comparison}\label{sec:few}

To answer \textbf{Q1}, we compare \our against seven baselines and summarize the results in Table \ref{tab:few-shot}. Overall, \our outperforms all baselines on all datasets. Compared with the general representation learning method, \ie, DeepWalk, \our achieve over $20$\% AUC improvement proving that the constructed structural-temporal subgraphs capture the dynamics of graph. For unsupervised anomaly detection methods, TADDY is the strongest baseline due to the Transformer-GNN encoder. However, it is also inferior to \our which can be attributed to the generalization power of LLMs. As to the semi-supervised methods, such as GDN and SAD, \our demonstrates notable improvements. For example, the relative AUC value improvements on the UCI Message dataset for different anomaly type in the 5-shot setting are $19\%$, $18.5\%$ and $20.3\%$, respectively. This is because \our employ ICL to excel the useful information of few labeled data. 

For different anomaly types, \our achieves stable improvements on CDA, LPL and HHL. We pretrain the dynamic-aware graph encoder for each dataset and detect different types of anomaly by only replacing the embedding of few labeled anomaly edges of the ICL template. As shown in Table~\ref{tab:few-shot}, the AUC of different anomaly types are over $80$\% indicating that \our is anomaly type-agnostic. Moreover, with the increase of labeled samples,  the performances of both \our and baseline methods are improved steadily. For example, compared to the 10-shot setting, the performance of SAD in the $1$-shot setting significantly decreased, with their AUC dropping by approximately $14$\%. This is because SAD is designed to detect anomaly with hundreds of labeled data. Conversely, \our still achieves over $0.82$ AUC on both two datasets. We attribute this to the effectiveness of ICL module which excites the advanced capabilities of LLMs.

\subsection{Efficiency Experiments}\label{sec:efficiency}
We study the efficiency of alignment and inference time to answer \textbf{Q2} and prove that \our is flexible for different LLM backbones. 

\begin{figure}[t]
    \centering
    \includegraphics[width=0.48\textwidth]{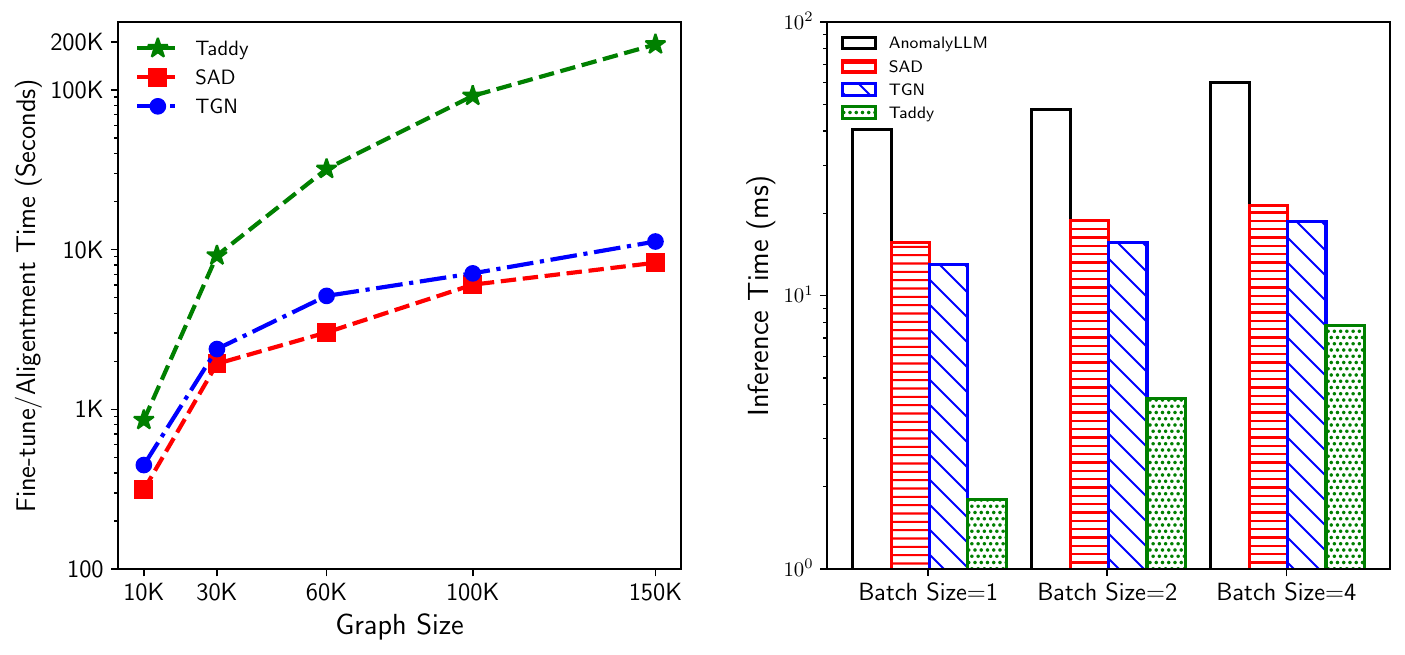}
    \vspace{-3ex}
    \caption{Inference time of \our}
    \label{fig:efficiency}
    \vspace{-3ex}
\end{figure}

For the compared baselines, the fine-tuning procedure need be conducted in few-shot anomaly detection. As shown in the left part of Figure~\ref{fig:efficiency}, the fine-tuning time increases linearly according to the number of edge sizes. For example, in $10$-shot anomaly detection of $60,000$ edges, the fine-tuning time of Taddy is over 10,000 seconds. As to \our, there has no fine-tuning procedure in few-shot anomaly detection. We can obtain the detection results of new anomaly types by only replacing the embedding of labeled edges in ICL template. The inference time of ICL detection is shown in the right part of Figure~\ref{fig:efficiency}. We can observe that the inference time of \our is comparable with other baselines under different batch sizes. This is because of the causal attention mechanism of LLMs. In model inference, the embeddings of the front part of ICL template stay unchanged for different input edges. Thus, \our is efficient for model inference and fine-tuning free for few-shot anomaly detection.

Furthermore, we study the alignment time that utilizes the pseudo label on BlogCatalog dataset to align the semantics of the neural language to dynamic graphs. As shown in Table~\ref{tab:alignment_time}, we count the alignment time of each epoch training by $30000$ pseudo label edges. In our experience, the alignment procedure would be convergence in $5$ epoch for different LLM backbones. As illustrated, the total alignment time of $30,000$ edges is about $1200$ seconds, which is acceptable for replacing the LLM backbone. Therefore, \our is simple and efficient to be updated with more powerful LLMs.
\begin{table}[ht]
\centering
\caption{ Alignment Fine-tuning Time of \our.}
\vspace{-3ex}
\label{tab:alignment_time}
\begin{tabular}{cc}
\hline
\textbf{Pseudo Label Edges} & \textbf{Alignment Time per Epoch (Seconds)} \\ \hline
10,000                      & 76.2                                        \\
30,000                      & 250.7                                       \\
100,000                     & 801.2                                       \\
150,000                     & 1203.2                                      \\ \hline
\end{tabular}
\vspace{-3ex}
\end{table}

\subsection{Ablation Results:}\label{appen:ablation}
To address \textbf{Q3}, we compare \our with three ablations to analyze the effectiveness of the proposed components. We remove the proposed dynamic-aware encoder, the alignment training module and the ICL detection respectively, and obtain w/o encoder,  w/o ICL and  w/o ICL.

\begin{table}[!htbp]
    \centering
    \caption{Ablation Results}\label{tab:ablate}
    \vspace{-3ex}
    \begin{tabular}{c|c|ccc}
    \hline
     &  & \multicolumn{3}{c}{\textbf{Anomaly Types}}  \\
    \multirow{-2}{*}{Dataset} & \multirow{-2}{*}{Method} & CDA & LPL & HHL \\ \hline
    \multicolumn{1}{c|}{} & w/o ICL & 0.7406 & 0.7465 & 0.7328  \\
    \multicolumn{1}{c|}{UCI} & w/o alignment  & 0.7849 & 0.7892 & 0.7994 \\
    \multicolumn{1}{c|}{Message} & w/o encoder       & 0.7727 & 0.7883 & 0.7822  \\
    \multicolumn{1}{c|}{} & \textbf{\our} & 0.8402 & 0.8456 & 0.8447 \\
    \hline
    & w/o ICL   & 0.7398 & 0.7421 & 0.7396  \\
    & w/o alignment  & 0.7767 & 0.7812 & 0.7726  \\
    & w/o encoder          & 0.7821 & 0.7726 & 0.7732 \\
    \multirow{-4}{*}{BlogCatalog} & \textbf{\our} & 0.8488 & 0.8546 & 0.8442 \\ \hline
    \end{tabular}
    \vspace{-3ex}
\end{table}
The experiment is conducted on BlogCatalog dataset and the results are shown in Table \ref{tab:ablate}. We observe: (1) Comparing the results of \our with w/o encoder, we observe the edge construction by focusing on subgraph embeddings from both sides can extract useful information and capture the evolving properties of edges in dynamic graphs. For example, the AUC improves from $0.7726$ to $0.8546$ on UCI Message dataset. (2) From the results of w/o ICL and \our, we can conclude that the ICL's capacity to efficiently utilize minimal labeled data is more effective than fine-tuning. (3) \our achieves the best performance compared to all ablations, which proves the effectiveness of the proposed techniques.

\begin{figure}[t]
    \centering
    \includegraphics[width=0.48\textwidth]{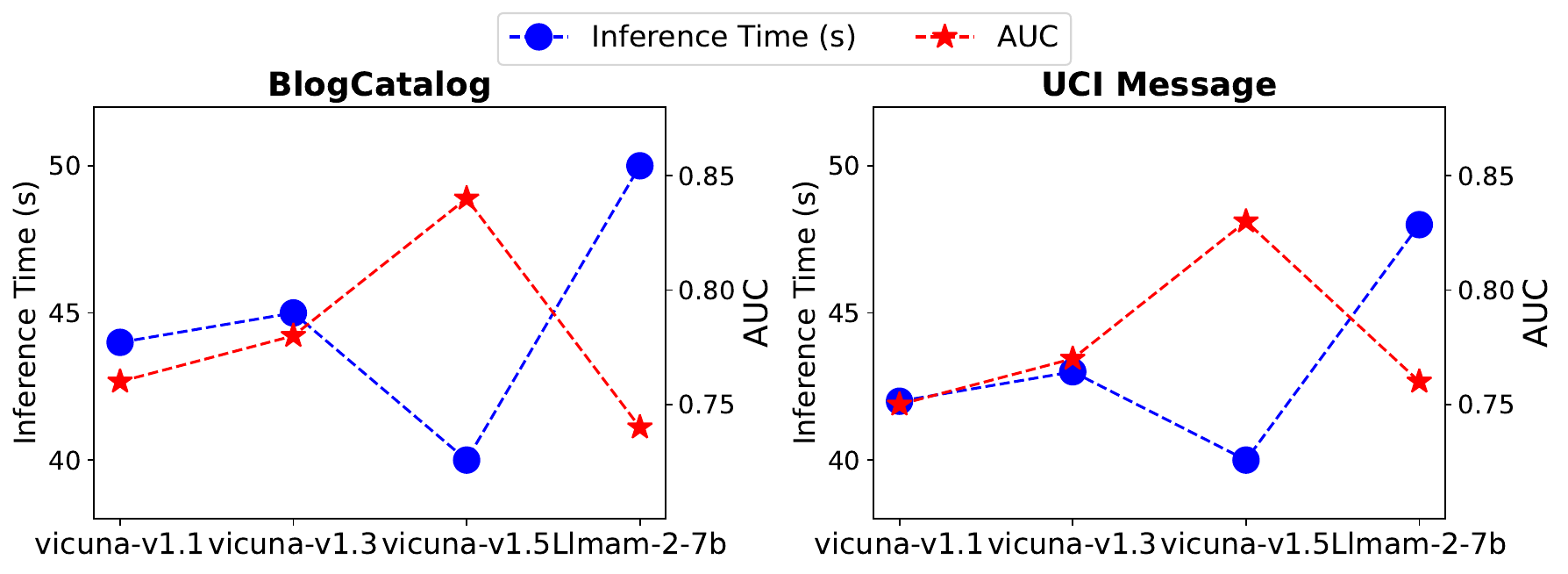}
    \vspace{-3ex}
    \caption{Performance of different LLM backbones}
    \label{fig:backbone}
    \vspace{-3ex}
\end{figure}

Moreover, we also explore the performance of \our under different LLM backbones on BlogCatalog and UCI Message datasets. As illustrated in Figure ~\ref{fig:backbone}, we assess the inference speed and AUC of various LLMs, including Llama-2-7B, vicuna-7B-v1.1,  vicuna-7B-v1.3 and vicuna-7B-v1.5. We can observe that vicuna-7B-v1.5 achieves the best performance and has the fastest inference time. To balance the performance and efficiency, we choose vicuna-7B-v1.5 as the LLM backbone.

\subsection{Performance on Real-World Labeled Dataset}
To answer \textbf{Q4}, we verify the performance of \our on two real-world datasets, \ie, T-Finance and T-Social, which have over 100 million edges. The results are summarized in Table \ref{tab:Real}. Overall, \our outperforms all baselines on all datasets. Compared with the state-of-the-art supervised learning method, \ie, TGN\cite{xu2020inductive}, \our achieve over $20.6$\% AUC improvement. For semi-supervised methods, \ie, GDN and SAD, \our demonstrates notable improvements. For example, compared to SAD, the relative AUC value improvements on the T-Social dataset for different shot settings are $21\%$, $20.7\%$ and $18.8\%$, respectively. These results indicate \our is potential to be used in large-scale dynamic graphs.
\begin{table}[htbp]
    \centering
 \caption{Performance on Real-World Labeled Dataset}\label{tab:Real}
    \vspace{-3ex}
    \begin{tabular}{c|c|ccc}
    \hline
    {Dataset} & {Method} & 1-shot & 5-shot & 10-shot \\ \hline
    \multicolumn{1}{c|}{} & AddGraph & 0.6126 & 0.6149 & 0.6277  \\
    \multicolumn{1}{c|}{} & TGN & 0.6646 & 0.6701 & 0.6865  \\
    \multicolumn{1}{c|}{} & GDN & 0.6672 & 0.6689 & 0.6898  \\
    \multicolumn{1}{c|}{\multirow{-4}{*}{T-Finance}} & SAD  & 0.6724 & 0.6754 & 0.6876 \\
    \multicolumn{1}{c|}{} & \textbf{\our} & \textbf{0.8018} & \textbf{0.8056} & \textbf{0.8087} \\ \hline
    & AddGraph  & 0.6116 & 0.6245 & 0.6221  \\
    & TGN  & 0.6706 & 0.6754 & 0.6887  \\
    & GDN  & 0.6694 & 0.6782 & 0.6908  \\
    \multirow{-4}{*}{T-Social} & SAD  & 0.6779 & 0.6746 & 0.6805  \\
    & \textbf{\our} & \textbf{0.8101} & \textbf{0.8187} & \textbf{0.8206} \\ \hline
    \end{tabular}
    \vspace{-3ex}
\end{table} 

\section{Conclusion}
In this paper, we are the first to integrate LLMs with dynamic graph anomaly detection, addressing the challenge of few-shot anomaly edge detection. \our leverages LLMs to effectively understand and represent the evolving relationships in dynamic graphs. We introduce a novel approach that reprograms the edge embedding to align the semantics between dynamic graph and LLMs. Moreover, an ICL strategy is designed to enable efficient and accurate detection of various anomaly types with a few labeled samples.   Extensive experiments across multiple datasets demonstrate that \our not only significantly outperforms existing methods in few-shot settings but also sets a new benchmark in the field.
\bibliographystyle{ACM-Reference-Format}
\bibliography{references}


\begin{thebibliography}{48}


\ifx \showCODEN    \undefined \def \showCODEN     #1{\unskip}     \fi
\ifx \showDOI      \undefined \def \showDOI       #1{#1}\fi
\ifx \showISBNx    \undefined \def \showISBNx     #1{\unskip}     \fi
\ifx \showISBNxiii \undefined \def \showISBNxiii  #1{\unskip}     \fi
\ifx \showISSN     \undefined \def \showISSN      #1{\unskip}     \fi
\ifx \showLCCN     \undefined \def \showLCCN      #1{\unskip}     \fi
\ifx \shownote     \undefined \def \shownote      #1{#1}          \fi
\ifx \showarticletitle \undefined \def \showarticletitle #1{#1}   \fi
\ifx \showURL      \undefined \def \showURL       {\relax}        \fi
\providecommand\bibfield[2]{#2}
\providecommand\bibinfo[2]{#2}
\providecommand\natexlab[1]{#1}
\providecommand\showeprint[2][]{arXiv:#2}

\bibitem[Balzarotti et~al\mbox{.}(2008)]%
        {balzarotti2008saner}
\bibfield{author}{\bibinfo{person}{Davide Balzarotti}, \bibinfo{person}{Marco
  Cova}, \bibinfo{person}{Vika Felmetsger}, \bibinfo{person}{Nenad Jovanovic},
  \bibinfo{person}{Engin Kirda}, \bibinfo{person}{Christopher Kruegel}, {and}
  \bibinfo{person}{Giovanni Vigna}.} \bibinfo{year}{2008}\natexlab{}.
\newblock \showarticletitle{Saner: Composing static and dynamic analysis to
  validate sanitization in web applications}. In \bibinfo{booktitle}{\emph{2008
  IEEE Symposium on Security and Privacy (sp 2008)}}. IEEE,
  \bibinfo{pages}{387--401}.
\newblock


\bibitem[Cai et~al\mbox{.}(2021)]%
        {cai2021structural}
\bibfield{author}{\bibinfo{person}{Lei Cai}, \bibinfo{person}{Zhengzhang Chen},
  \bibinfo{person}{Chen Luo}, \bibinfo{person}{Jiaping Gui},
  \bibinfo{person}{Jingchao Ni}, \bibinfo{person}{Ding Li}, {and}
  \bibinfo{person}{Haifeng Chen}.} \bibinfo{year}{2021}\natexlab{}.
\newblock \showarticletitle{Structural temporal graph neural networks for
  anomaly detection in dynamic graphs}. In
  \bibinfo{booktitle}{\emph{Proceedings of the 30th ACM international
  conference on Information \& Knowledge Management}}.
  \bibinfo{pages}{3747--3756}.
\newblock


\bibitem[Deng et~al\mbox{.}(2019)]%
        {deng2019learning}
\bibfield{author}{\bibinfo{person}{Songgaojun Deng}, \bibinfo{person}{Huzefa
  Rangwala}, {and} \bibinfo{person}{Yue Ning}.}
  \bibinfo{year}{2019}\natexlab{}.
\newblock \showarticletitle{Learning dynamic context graphs for predicting
  social events}. In \bibinfo{booktitle}{\emph{Proceedings of the 25th ACM
  SIGKDD International Conference on Knowledge Discovery \& Data Mining}}.
  \bibinfo{pages}{1007--1016}.
\newblock


\bibitem[Ding et~al\mbox{.}(2019)]%
        {ding2019deep}
\bibfield{author}{\bibinfo{person}{Kaize Ding}, \bibinfo{person}{Jundong Li},
  \bibinfo{person}{Rohit Bhanushali}, {and} \bibinfo{person}{Huan Liu}.}
  \bibinfo{year}{2019}\natexlab{}.
\newblock \showarticletitle{Deep anomaly detection on attributed networks}. In
  \bibinfo{booktitle}{\emph{Proceedings of the 2019 SIAM International
  Conference on Data Mining}}. SIAM, \bibinfo{pages}{594--602}.
\newblock


\bibitem[Ding et~al\mbox{.}(2020)]%
        {ding2020graph}
\bibfield{author}{\bibinfo{person}{Kaize Ding}, \bibinfo{person}{Jianling
  Wang}, \bibinfo{person}{Jundong Li}, \bibinfo{person}{Kai Shu},
  \bibinfo{person}{Chenghao Liu}, {and} \bibinfo{person}{Huan Liu}.}
  \bibinfo{year}{2020}\natexlab{}.
\newblock \showarticletitle{Graph prototypical networks for few-shot learning
  on attributed networks}. In \bibinfo{booktitle}{\emph{Proceedings of the 29th
  ACM International Conference on Information \& Knowledge Management}}.
  \bibinfo{pages}{295--304}.
\newblock


\bibitem[Ding et~al\mbox{.}(2021)]%
        {ding2021few}
\bibfield{author}{\bibinfo{person}{Kaize Ding}, \bibinfo{person}{Qinghai Zhou},
  \bibinfo{person}{Hanghang Tong}, {and} \bibinfo{person}{Huan Liu}.}
  \bibinfo{year}{2021}\natexlab{}.
\newblock \showarticletitle{Few-shot network anomaly detection via
  cross-network meta-learning}. In \bibinfo{booktitle}{\emph{Proceedings of the
  Web Conference 2021}}. \bibinfo{pages}{2448--2456}.
\newblock


\bibitem[Duan et~al\mbox{.}(2020a)]%
        {duan2020aane}
\bibfield{author}{\bibinfo{person}{Dongsheng Duan}, \bibinfo{person}{Lingling
  Tong}, \bibinfo{person}{Yangxi Li}, \bibinfo{person}{Jie Lu},
  \bibinfo{person}{Lei Shi}, {and} \bibinfo{person}{Cheng Zhang}.}
  \bibinfo{year}{2020}\natexlab{a}.
\newblock \showarticletitle{Aane: Anomaly aware network embedding for anomalous
  link detection}. In \bibinfo{booktitle}{\emph{2020 IEEE International
  Conference on Data Mining (ICDM)}}. IEEE, \bibinfo{pages}{1002--1007}.
\newblock


\bibitem[Duan et~al\mbox{.}(2020b)]%
        {DuanTLLSZ20}
\bibfield{author}{\bibinfo{person}{Dongsheng Duan}, \bibinfo{person}{Lingling
  Tong}, \bibinfo{person}{Yangxi Li}, \bibinfo{person}{Jie Lu},
  \bibinfo{person}{Lei Shi}, {and} \bibinfo{person}{Cheng Zhang}.}
  \bibinfo{year}{2020}\natexlab{b}.
\newblock \showarticletitle{{AANE:} Anomaly Aware Network Embedding For
  Anomalous Link Detection}. In \bibinfo{booktitle}{\emph{20th {IEEE}
  International Conference on Data Mining, {ICDM}}}.
  \bibinfo{pages}{1002--1007}.
\newblock


\bibitem[Fifty et~al\mbox{.}(2023)]%
        {fifty2023context}
\bibfield{author}{\bibinfo{person}{Christopher Fifty}, \bibinfo{person}{Jure
  Leskovec}, {and} \bibinfo{person}{Sebastian Thrun}.}
  \bibinfo{year}{2023}\natexlab{}.
\newblock \showarticletitle{In-Context Learning for Few-Shot Molecular Property
  Prediction}.
\newblock \bibinfo{journal}{\emph{arXiv preprint arXiv:2310.08863}}
  (\bibinfo{year}{2023}).
\newblock


\bibitem[Grover and Leskovec(2016)]%
        {GroverL16}
\bibfield{author}{\bibinfo{person}{Aditya Grover} {and} \bibinfo{person}{Jure
  Leskovec}.} \bibinfo{year}{2016}\natexlab{}.
\newblock \showarticletitle{node2vec: Scalable Feature Learning for Networks}.
  In \bibinfo{booktitle}{\emph{Proceedings of the 22nd ACM SIGKDD Conference on
  Knowledge Discovery and Data Mining}}. \bibinfo{pages}{855--864}.
\newblock


\bibitem[Guo et~al\mbox{.}(2021)]%
        {guo2021few}
\bibfield{author}{\bibinfo{person}{Zhichun Guo}, \bibinfo{person}{Chuxu Zhang},
  \bibinfo{person}{Wenhao Yu}, \bibinfo{person}{John Herr},
  \bibinfo{person}{Olaf Wiest}, \bibinfo{person}{Meng Jiang}, {and}
  \bibinfo{person}{Nitesh~V Chawla}.} \bibinfo{year}{2021}\natexlab{}.
\newblock \showarticletitle{Few-shot graph learning for molecular property
  prediction}. In \bibinfo{booktitle}{\emph{Proceedings of the web conference
  2021}}. \bibinfo{pages}{2559--2567}.
\newblock


\bibitem[Hao et~al\mbox{.}(2023)]%
        {hao2023reasoning}
\bibfield{author}{\bibinfo{person}{Shibo Hao}, \bibinfo{person}{Yi Gu},
  \bibinfo{person}{Haodi Ma}, \bibinfo{person}{Joshua~Jiahua Hong},
  \bibinfo{person}{Zhen Wang}, \bibinfo{person}{Daisy~Zhe Wang}, {and}
  \bibinfo{person}{Zhiting Hu}.} \bibinfo{year}{2023}\natexlab{}.
\newblock \showarticletitle{Reasoning with language model is planning with
  world model}.
\newblock \bibinfo{journal}{\emph{arXiv preprint arXiv:2305.14992}}
  (\bibinfo{year}{2023}).
\newblock


\bibitem[Huang et~al\mbox{.}(2022)]%
        {huang2022dgraph}
\bibfield{author}{\bibinfo{person}{Xuanwen Huang}, \bibinfo{person}{Yang Yang},
  \bibinfo{person}{Yang Wang}, \bibinfo{person}{Chunping Wang},
  \bibinfo{person}{Zhisheng Zhang}, \bibinfo{person}{Jiarong Xu},
  \bibinfo{person}{Lei Chen}, {and} \bibinfo{person}{Michalis Vazirgiannis}.}
  \bibinfo{year}{2022}\natexlab{}.
\newblock \showarticletitle{Dgraph: A large-scale financial dataset for graph
  anomaly detection}.
\newblock \bibinfo{journal}{\emph{Advances in Neural Information Processing
  Systems}}  \bibinfo{volume}{35} (\bibinfo{year}{2022}),
  \bibinfo{pages}{22765--22777}.
\newblock


\bibitem[Jin et~al\mbox{.}(2023)]%
        {jin2023time}
\bibfield{author}{\bibinfo{person}{Ming Jin}, \bibinfo{person}{Shiyu Wang},
  \bibinfo{person}{Lintao Ma}, \bibinfo{person}{Zhixuan Chu},
  \bibinfo{person}{James~Y Zhang}, \bibinfo{person}{Xiaoming Shi},
  \bibinfo{person}{Pin-Yu Chen}, \bibinfo{person}{Yuxuan Liang},
  \bibinfo{person}{Yuan-Fang Li}, \bibinfo{person}{Shirui Pan},
  {et~al\mbox{.}}} \bibinfo{year}{2023}\natexlab{}.
\newblock \showarticletitle{Time-llm: Time series forecasting by reprogramming
  large language models}.
\newblock \bibinfo{journal}{\emph{arXiv preprint arXiv:2310.01728}}
  (\bibinfo{year}{2023}).
\newblock


\bibitem[John et~al\mbox{.}(2017)]%
        {john2017service}
\bibfield{author}{\bibinfo{person}{Wolfgang John}, \bibinfo{person}{Guido
  Marchetto}, \bibinfo{person}{Felici{\'a}n N{\'e}meth},
  \bibinfo{person}{Pontus Skoldstrom}, \bibinfo{person}{Rebecca Steinert},
  \bibinfo{person}{Catalin Meirosu}, \bibinfo{person}{Ioanna Papafili}, {and}
  \bibinfo{person}{Kostas Pentikousis}.} \bibinfo{year}{2017}\natexlab{}.
\newblock \showarticletitle{Service provider devops}.
\newblock \bibinfo{journal}{\emph{IEEE Communications Magazine}}
  \bibinfo{volume}{55}, \bibinfo{number}{1} (\bibinfo{year}{2017}),
  \bibinfo{pages}{204--211}.
\newblock


\bibitem[Kondor and Lafferty(2002)]%
        {KondorL02}
\bibfield{author}{\bibinfo{person}{Risi Kondor} {and} \bibinfo{person}{John~D.
  Lafferty}.} \bibinfo{year}{2002}\natexlab{}.
\newblock \showarticletitle{Diffusion Kernels on Graphs and Other Discrete
  Input Spaces}. In \bibinfo{booktitle}{\emph{Machine Learning, Proceedings of
  the Nineteenth International Conference (ICML 2002)}}.
  \bibinfo{pages}{315--322}.
\newblock


\bibitem[Li et~al\mbox{.}(2020)]%
        {li2020few}
\bibfield{author}{\bibinfo{person}{Ruirui Li}, \bibinfo{person}{Xian Wu},
  \bibinfo{person}{Xian Wu}, {and} \bibinfo{person}{Wei Wang}.}
  \bibinfo{year}{2020}\natexlab{}.
\newblock \showarticletitle{Few-shot learning for new user recommendation in
  location-based social networks}. In \bibinfo{booktitle}{\emph{Proceedings of
  The Web Conference 2020}}. \bibinfo{pages}{2472--2478}.
\newblock


\bibitem[Liu et~al\mbox{.}(2022)]%
        {liu2022deep}
\bibfield{author}{\bibinfo{person}{Jiaying Liu}, \bibinfo{person}{Feng Xia},
  \bibinfo{person}{Xu Feng}, \bibinfo{person}{Jing Ren}, {and}
  \bibinfo{person}{Huan Liu}.} \bibinfo{year}{2022}\natexlab{}.
\newblock \showarticletitle{Deep graph learning for anomalous citation
  detection}.
\newblock \bibinfo{journal}{\emph{IEEE Transactions on Neural Networks and
  Learning Systems}} \bibinfo{volume}{33}, \bibinfo{number}{6}
  (\bibinfo{year}{2022}), \bibinfo{pages}{2543--2557}.
\newblock


\bibitem[Liu et~al\mbox{.}(2021b)]%
        {liu2021anomaly}
\bibfield{author}{\bibinfo{person}{Yixin Liu}, \bibinfo{person}{Shirui Pan},
  \bibinfo{person}{Yu~Guang Wang}, \bibinfo{person}{Fei Xiong},
  \bibinfo{person}{Liang Wang}, \bibinfo{person}{Qingfeng Chen}, {and}
  \bibinfo{person}{Vincent~CS Lee}.} \bibinfo{year}{2021}\natexlab{b}.
\newblock \showarticletitle{Anomaly detection in dynamic graphs via
  transformer}.
\newblock \bibinfo{journal}{\emph{IEEE Transactions on Knowledge and Data
  Engineering}} (\bibinfo{year}{2021}).
\newblock


\bibitem[Liu et~al\mbox{.}(2021a)]%
        {liu2021relative}
\bibfield{author}{\bibinfo{person}{Zemin Liu}, \bibinfo{person}{Yuan Fang},
  \bibinfo{person}{Chenghao Liu}, {and} \bibinfo{person}{Steven~CH Hoi}.}
  \bibinfo{year}{2021}\natexlab{a}.
\newblock \showarticletitle{Relative and absolute location embedding for
  few-shot node classification on graph}. In
  \bibinfo{booktitle}{\emph{Proceedings of the AAAI conference on artificial
  intelligence}}, Vol.~\bibinfo{volume}{35}. \bibinfo{pages}{4267--4275}.
\newblock


\bibitem[Liu et~al\mbox{.}(2023)]%
        {liu2023rgse}
\bibfield{author}{\bibinfo{person}{Zhen Liu}, \bibinfo{person}{Wenbo Zuo},
  \bibinfo{person}{Dongning Zhang}, {and} \bibinfo{person}{Xiaodong Feng}.}
  \bibinfo{year}{2023}\natexlab{}.
\newblock \showarticletitle{RGSE: Robust Graph Structure Embedding for
  Anomalous Link Detection}.
\newblock \bibinfo{journal}{\emph{IEEE Transactions on Big Data}}
  (\bibinfo{year}{2023}).
\newblock


\bibitem[Lu et~al\mbox{.}(2022)]%
        {lu2022bright}
\bibfield{author}{\bibinfo{person}{Mingxuan Lu}, \bibinfo{person}{Zhichao Han},
  \bibinfo{person}{Susie~Xi Rao}, \bibinfo{person}{Zitao Zhang},
  \bibinfo{person}{Yang Zhao}, \bibinfo{person}{Yinan Shan},
  \bibinfo{person}{Ramesh Raghunathan}, \bibinfo{person}{Ce Zhang}, {and}
  \bibinfo{person}{Jiawei Jiang}.} \bibinfo{year}{2022}\natexlab{}.
\newblock \showarticletitle{BRIGHT-Graph Neural Networks in Real-Time Fraud
  Detection}. In \bibinfo{booktitle}{\emph{Proceedings of the 31st ACM
  International Conference on Information \& Knowledge Management}}.
  \bibinfo{pages}{3342--3351}.
\newblock


\bibitem[Ma et~al\mbox{.}(2021)]%
        {ma2021comprehensive}
\bibfield{author}{\bibinfo{person}{Xiaoxiao Ma}, \bibinfo{person}{Jia Wu},
  \bibinfo{person}{Shan Xue}, \bibinfo{person}{Jian Yang},
  \bibinfo{person}{Chuan Zhou}, \bibinfo{person}{Quan~Z Sheng},
  \bibinfo{person}{Hui Xiong}, {and} \bibinfo{person}{Leman Akoglu}.}
  \bibinfo{year}{2021}\natexlab{}.
\newblock \showarticletitle{A comprehensive survey on graph anomaly detection
  with deep learning}.
\newblock \bibinfo{journal}{\emph{IEEE Transactions on Knowledge and Data
  Engineering}} (\bibinfo{year}{2021}).
\newblock


\bibitem[Meng et~al\mbox{.}(2021)]%
        {meng2021semi}
\bibfield{author}{\bibinfo{person}{Xuying Meng}, \bibinfo{person}{Suhang Wang},
  \bibinfo{person}{Zhimin Liang}, \bibinfo{person}{Di Yao},
  \bibinfo{person}{Jihua Zhou}, {and} \bibinfo{person}{Yujun Zhang}.}
  \bibinfo{year}{2021}\natexlab{}.
\newblock \showarticletitle{Semi-supervised anomaly detection in dynamic
  communication networks}.
\newblock \bibinfo{journal}{\emph{Information Sciences}}  \bibinfo{volume}{571}
  (\bibinfo{year}{2021}), \bibinfo{pages}{527--542}.
\newblock


\bibitem[Miz et~al\mbox{.}(2019)]%
        {miz2019anomaly}
\bibfield{author}{\bibinfo{person}{Volodymyr Miz}, \bibinfo{person}{Benjamin
  Ricaud}, \bibinfo{person}{Kirell Benzi}, {and} \bibinfo{person}{Pierre
  Vandergheynst}.} \bibinfo{year}{2019}\natexlab{}.
\newblock \showarticletitle{Anomaly detection in the dynamics of web and social
  networks using associative memory}. In \bibinfo{booktitle}{\emph{The World
  Wide Web Conference}}. \bibinfo{pages}{1290--1299}.
\newblock


\bibitem[Opsahl and Panzarasa(2009)]%
        {opsahl2009clustering}
\bibfield{author}{\bibinfo{person}{Tore Opsahl} {and} \bibinfo{person}{Pietro
  Panzarasa}.} \bibinfo{year}{2009}\natexlab{}.
\newblock \showarticletitle{Clustering in weighted networks}.
\newblock \bibinfo{journal}{\emph{Social networks}} \bibinfo{volume}{31},
  \bibinfo{number}{2} (\bibinfo{year}{2009}), \bibinfo{pages}{155--163}.
\newblock


\bibitem[Page et~al\mbox{.}(1998)]%
        {page1998pagerank}
\bibfield{author}{\bibinfo{person}{Lawrence Page}, \bibinfo{person}{Sergey
  Brin}, \bibinfo{person}{Rajeev Motwani}, {and} \bibinfo{person}{Terry
  Winograd}.} \bibinfo{year}{1998}\natexlab{}.
\newblock \bibinfo{booktitle}{\emph{The pagerank citation ranking: Bring order
  to the web}}.
\newblock \bibinfo{type}{{T}echnical {R}eport}. \bibinfo{institution}{Technical
  report, stanford University}.
\newblock


\bibitem[Pawelczyk et~al\mbox{.}(2023)]%
        {pawelczyk2023context}
\bibfield{author}{\bibinfo{person}{Martin Pawelczyk}, \bibinfo{person}{Seth
  Neel}, {and} \bibinfo{person}{Himabindu Lakkaraju}.}
  \bibinfo{year}{2023}\natexlab{}.
\newblock \showarticletitle{In-context unlearning: Language models as few shot
  unlearners}.
\newblock \bibinfo{journal}{\emph{arXiv preprint arXiv:2310.07579}}
  (\bibinfo{year}{2023}).
\newblock


\bibitem[Perozzi et~al\mbox{.}(2014)]%
        {perozzi2014deepwalk}
\bibfield{author}{\bibinfo{person}{Bryan Perozzi}, \bibinfo{person}{Rami
  Al-Rfou}, {and} \bibinfo{person}{Steven Skiena}.}
  \bibinfo{year}{2014}\natexlab{}.
\newblock \showarticletitle{Deepwalk: Online learning of social
  representations}. In \bibinfo{booktitle}{\emph{Proceedings of the 20th ACM
  SIGKDD international conference on Knowledge discovery and data mining}}.
  \bibinfo{pages}{701--710}.
\newblock


\bibitem[Ranshous et~al\mbox{.}(2016)]%
        {ranshous2016scalable}
\bibfield{author}{\bibinfo{person}{Stephen Ranshous}, \bibinfo{person}{Steve
  Harenberg}, \bibinfo{person}{Kshitij Sharma}, {and} \bibinfo{person}{Nagiza~F
  Samatova}.} \bibinfo{year}{2016}\natexlab{}.
\newblock \showarticletitle{A scalable approach for outlier detection in edge
  streams using sketch-based approximations}. In
  \bibinfo{booktitle}{\emph{Proceedings of the 2016 SIAM international
  conference on data mining}}. SIAM, \bibinfo{pages}{189--197}.
\newblock


\bibitem[Sun et~al\mbox{.}(2023)]%
        {sun2023all}
\bibfield{author}{\bibinfo{person}{Xiangguo Sun}, \bibinfo{person}{Hong Cheng},
  \bibinfo{person}{Jia Li}, \bibinfo{person}{Bo Liu}, {and}
  \bibinfo{person}{Jihong Guan}.} \bibinfo{year}{2023}\natexlab{}.
\newblock \showarticletitle{All in One: Multi-Task Prompting for Graph Neural
  Networks}.
\newblock  (\bibinfo{year}{2023}).
\newblock


\bibitem[Tang et~al\mbox{.}(2022)]%
        {tang2022rethinking}
\bibfield{author}{\bibinfo{person}{Jianheng Tang}, \bibinfo{person}{Jiajin Li},
  \bibinfo{person}{Ziqi Gao}, {and} \bibinfo{person}{Jia Li}.}
  \bibinfo{year}{2022}\natexlab{}.
\newblock \showarticletitle{Rethinking graph neural networks for anomaly
  detection}. In \bibinfo{booktitle}{\emph{International Conference on Machine
  Learning}}. PMLR, \bibinfo{pages}{21076--21089}.
\newblock


\bibitem[Tang et~al\mbox{.}(2023)]%
        {tang2023graphgpt}
\bibfield{author}{\bibinfo{person}{Jiabin Tang}, \bibinfo{person}{Yuhao Yang},
  \bibinfo{person}{Wei Wei}, \bibinfo{person}{Lei Shi}, \bibinfo{person}{Lixin
  Su}, \bibinfo{person}{Suqi Cheng}, \bibinfo{person}{Dawei Yin}, {and}
  \bibinfo{person}{Chao Huang}.} \bibinfo{year}{2023}\natexlab{}.
\newblock \showarticletitle{Graphgpt: Graph instruction tuning for large
  language models}.
\newblock \bibinfo{journal}{\emph{arXiv preprint arXiv:2310.13023}}
  (\bibinfo{year}{2023}).
\newblock


\bibitem[Tang and Liu(2009)]%
        {tang2009relational}
\bibfield{author}{\bibinfo{person}{Lei Tang} {and} \bibinfo{person}{Huan Liu}.}
  \bibinfo{year}{2009}\natexlab{}.
\newblock \showarticletitle{Relational learning via latent social dimensions}.
  In \bibinfo{booktitle}{\emph{Proceedings of the 15th ACM SIGKDD international
  conference on Knowledge discovery and data mining}}.
  \bibinfo{pages}{817--826}.
\newblock


\bibitem[Tian et~al\mbox{.}(2023)]%
        {tian2023sad}
\bibfield{author}{\bibinfo{person}{Sheng Tian}, \bibinfo{person}{Jihai Dong},
  \bibinfo{person}{Jintang Li}, \bibinfo{person}{Wenlong Zhao},
  \bibinfo{person}{Xiaolong Xu}, \bibinfo{person}{Bowen Song},
  \bibinfo{person}{Changhua Meng}, \bibinfo{person}{Tianyi Zhang},
  \bibinfo{person}{Liang Chen}, {et~al\mbox{.}}}
  \bibinfo{year}{2023}\natexlab{}.
\newblock \showarticletitle{SAD: Semi-Supervised Anomaly Detection on Dynamic
  Graphs}.
\newblock \bibinfo{journal}{\emph{arXiv preprint arXiv:2305.13573}}
  (\bibinfo{year}{2023}).
\newblock


\bibitem[Wang et~al\mbox{.}(2019)]%
        {wang2019semi}
\bibfield{author}{\bibinfo{person}{Daixin Wang}, \bibinfo{person}{Jianbin Lin},
  \bibinfo{person}{Peng Cui}, \bibinfo{person}{Quanhui Jia},
  \bibinfo{person}{Zhen Wang}, \bibinfo{person}{Yanming Fang},
  \bibinfo{person}{Quan Yu}, \bibinfo{person}{Jun Zhou},
  \bibinfo{person}{Shuang Yang}, {and} \bibinfo{person}{Yuan Qi}.}
  \bibinfo{year}{2019}\natexlab{}.
\newblock \showarticletitle{A semi-supervised graph attentive network for
  financial fraud detection}. In \bibinfo{booktitle}{\emph{2019 IEEE
  International Conference on Data Mining (ICDM)}}. IEEE,
  \bibinfo{pages}{598--607}.
\newblock


\bibitem[Wang and Qiao(2019)]%
        {wang2019nodes}
\bibfield{author}{\bibinfo{person}{Huan Wang} {and} \bibinfo{person}{Chunming
  Qiao}.} \bibinfo{year}{2019}\natexlab{}.
\newblock \showarticletitle{A nodes' evolution diversity inspired method to
  detect anomalies in dynamic social networks}.
\newblock \bibinfo{journal}{\emph{IEEE Transactions on Knowledge and Data
  Engineering}} \bibinfo{volume}{32}, \bibinfo{number}{10}
  (\bibinfo{year}{2019}), \bibinfo{pages}{1868--1880}.
\newblock


\bibitem[Wei et~al\mbox{.}(2022)]%
        {wei2022contrastive}
\bibfield{author}{\bibinfo{person}{Wei Wei}, \bibinfo{person}{Chao Huang},
  \bibinfo{person}{Lianghao Xia}, \bibinfo{person}{Yong Xu},
  \bibinfo{person}{Jiashu Zhao}, {and} \bibinfo{person}{Dawei Yin}.}
  \bibinfo{year}{2022}\natexlab{}.
\newblock \showarticletitle{Contrastive meta learning with behavior
  multiplicity for recommendation}. In \bibinfo{booktitle}{\emph{Proceedings of
  the fifteenth ACM international conference on web search and data mining}}.
  \bibinfo{pages}{1120--1128}.
\newblock


\bibitem[Wi et~al\mbox{.}(2022)]%
        {wi2022hiddencpg}
\bibfield{author}{\bibinfo{person}{Seongil Wi}, \bibinfo{person}{Sijae Woo},
  \bibinfo{person}{Joyce~Jiyoung Whang}, {and} \bibinfo{person}{Sooel Son}.}
  \bibinfo{year}{2022}\natexlab{}.
\newblock \showarticletitle{HiddenCPG: large-scale vulnerable clone detection
  using subgraph isomorphism of code property graphs}. In
  \bibinfo{booktitle}{\emph{Proceedings of the ACM Web Conference 2022}}.
  \bibinfo{pages}{755--766}.
\newblock


\bibitem[Wu et~al\mbox{.}(2022)]%
        {wu2022graph}
\bibfield{author}{\bibinfo{person}{Shiwen Wu}, \bibinfo{person}{Fei Sun},
  \bibinfo{person}{Wentao Zhang}, \bibinfo{person}{Xu Xie}, {and}
  \bibinfo{person}{Bin Cui}.} \bibinfo{year}{2022}\natexlab{}.
\newblock \showarticletitle{Graph neural networks in recommender systems: a
  survey}.
\newblock \bibinfo{journal}{\emph{Comput. Surveys}} \bibinfo{volume}{55},
  \bibinfo{number}{5} (\bibinfo{year}{2022}), \bibinfo{pages}{1--37}.
\newblock


\bibitem[Xu et~al\mbox{.}(2020)]%
        {xu2020inductive}
\bibfield{author}{\bibinfo{person}{Da Xu}, \bibinfo{person}{Chuanwei Ruan},
  \bibinfo{person}{Evren Korpeoglu}, \bibinfo{person}{Sushant Kumar}, {and}
  \bibinfo{person}{Kannan Achan}.} \bibinfo{year}{2020}\natexlab{}.
\newblock \showarticletitle{Inductive representation learning on temporal
  graphs}.
\newblock \bibinfo{journal}{\emph{arXiv preprint arXiv:2002.07962}}
  (\bibinfo{year}{2020}).
\newblock


\bibitem[Xu et~al\mbox{.}(2023)]%
        {xu2023metagad}
\bibfield{author}{\bibinfo{person}{Xiongxiao Xu}, \bibinfo{person}{Kaize Ding},
  \bibinfo{person}{Canyu Chen}, {and} \bibinfo{person}{Kai Shu}.}
  \bibinfo{year}{2023}\natexlab{}.
\newblock \showarticletitle{MetaGAD: Learning to Meta Transfer for Few-shot
  Graph Anomaly Detection}.
\newblock \bibinfo{journal}{\emph{arXiv preprint arXiv:2305.10668}}
  (\bibinfo{year}{2023}).
\newblock


\bibitem[Yang et~al\mbox{.}(2020)]%
        {yang2020h}
\bibfield{author}{\bibinfo{person}{Chenming Yang}, \bibinfo{person}{Liang
  Zhou}, \bibinfo{person}{Hui Wen}, \bibinfo{person}{Zhiheng Zhou}, {and}
  \bibinfo{person}{Yue Wu}.} \bibinfo{year}{2020}\natexlab{}.
\newblock \showarticletitle{H-vgrae: A hierarchical stochastic spatial-temporal
  embedding method for robust anomaly detection in dynamic networks}.
\newblock \bibinfo{journal}{\emph{arXiv preprint arXiv:2007.06903}}
  (\bibinfo{year}{2020}).
\newblock


\bibitem[Ye et~al\mbox{.}(2023)]%
        {ye2023natural}
\bibfield{author}{\bibinfo{person}{Ruosong Ye}, \bibinfo{person}{Caiqi Zhang},
  \bibinfo{person}{Runhui Wang}, \bibinfo{person}{Shuyuan Xu}, {and}
  \bibinfo{person}{Yongfeng Zhang}.} \bibinfo{year}{2023}\natexlab{}.
\newblock \showarticletitle{Natural language is all a graph needs}.
\newblock \bibinfo{journal}{\emph{arXiv preprint arXiv:2308.07134}}
  (\bibinfo{year}{2023}).
\newblock


\bibitem[Yu et~al\mbox{.}(2018)]%
        {yu2018netwalk}
\bibfield{author}{\bibinfo{person}{Wenchao Yu}, \bibinfo{person}{Wei Cheng},
  \bibinfo{person}{Charu~C Aggarwal}, \bibinfo{person}{Kai Zhang},
  \bibinfo{person}{Haifeng Chen}, {and} \bibinfo{person}{Wei Wang}.}
  \bibinfo{year}{2018}\natexlab{}.
\newblock \showarticletitle{Netwalk: A flexible deep embedding approach for
  anomaly detection in dynamic networks}. In
  \bibinfo{booktitle}{\emph{Proceedings of the 24th ACM SIGKDD international
  conference on knowledge discovery \& data mining}}.
  \bibinfo{pages}{2672--2681}.
\newblock


\bibitem[Zhao et~al\mbox{.}(2023)]%
        {zhao2023doubleadapt}
\bibfield{author}{\bibinfo{person}{Lifan Zhao}, \bibinfo{person}{Shuming Kong},
  {and} \bibinfo{person}{Yanyan Shen}.} \bibinfo{year}{2023}\natexlab{}.
\newblock \showarticletitle{DoubleAdapt: A Meta-learning Approach to
  Incremental Learning for Stock Trend Forecasting}. In
  \bibinfo{booktitle}{\emph{Proceedings of the 29th ACM SIGKDD Conference on
  Knowledge Discovery and Data Mining}}. \bibinfo{pages}{3492--3503}.
\newblock


\bibitem[Zheng et~al\mbox{.}(2019)]%
        {zheng2019addgraph}
\bibfield{author}{\bibinfo{person}{Li Zheng}, \bibinfo{person}{Zhenpeng Li},
  \bibinfo{person}{Jian Li}, \bibinfo{person}{Zhao Li}, {and}
  \bibinfo{person}{Jun Gao}.} \bibinfo{year}{2019}\natexlab{}.
\newblock \showarticletitle{AddGraph: Anomaly Detection in Dynamic Graph Using
  Attention-based Temporal GCN.}. In \bibinfo{booktitle}{\emph{IJCAI}},
  Vol.~\bibinfo{volume}{3}. \bibinfo{pages}{7}.
\newblock


\bibitem[Zhu et~al\mbox{.}(2023)]%
        {zhu2023wingnn}
\bibfield{author}{\bibinfo{person}{Yifan Zhu}, \bibinfo{person}{Fangpeng Cong},
  \bibinfo{person}{Dan Zhang}, \bibinfo{person}{Wenwen Gong},
  \bibinfo{person}{Qika Lin}, \bibinfo{person}{Wenzheng Feng},
  \bibinfo{person}{Yuxiao Dong}, {and} \bibinfo{person}{Jie Tang}.}
  \bibinfo{year}{2023}\natexlab{}.
\newblock \showarticletitle{WinGNN: Dynamic Graph Neural Networks with Random
  Gradient Aggregation Window}. In \bibinfo{booktitle}{\emph{Proceedings of the
  29th ACM SIGKDD Conference on Knowledge Discovery and Data Mining}}.
  \bibinfo{pages}{3650--3662}.
\newblock


\end{thebibliography}
\clearpage
\appendix
\section{Appendix}

\subsection{Detail of Dynamic Encoder}\label{appen:detail}
\subsubsection{Calculation of Diffusion Matrix}
Given the adjacency matrix $\mathbf{A}^t \in \mathbb{R}^{n \times n}$ at timestamp $t$, we calculate the diffusion matrix $\mathbf{D}^t \in \mathbb{R}^{N \times N}$ to select related nodes for the target edge. For brevity, we ignore the superscript $t$, and the diffusion matrix $\mathbf{D}$ can be calculated according to the adjacency matrix $\mathbf{A}$:
$$\mathbf{D} = \sum_{m=0}^\infty \theta_m \mathbf{T}^m,$$
where $\mathbf{T} \in \mathbb{R}^{n \times n}$ is the generalized transition matrix and $\theta_m$ is the weighting coefficient indicating the ratio of global-local information. It requires that $\sum_{m=0}^\infty \theta_m =1, \theta_m \in [0,1]$ and the eigenvalues $\lambda_r$ of $\mathbf{T}$ are bounded by $\lambda_r \in [0, 1]$ to guarantee convergence. Different instantiations of diffusion matrix can be computed by applying specific definitions of $\mathbf{T}$ and $\theta$. For instance, Personalized PageRank (PPR)~\cite{page1998pagerank} chooses $\mathbf{T} = \mathbf{A}\mathbf{S}^{-1}$ and $\theta_m = \alpha(1-\alpha)^m$, where $\mathbf{S} \in \mathbb{R}^{n \times n}$ is the diagonal degree matrix and $\alpha \in (0, 1)$ is the teleport probability. Another popular example of diffusion matrix is the heat kernal~\cite{KondorL02}, which chooses $\mathbf{T}=\mathbf{A}\mathbf{S}^{-1}$ and $\theta_m = e^{-\beta}\beta^m/m!$, where $\beta$ is the diffusion time. The solutions to PPR and heat kernel can be formulated as:
$$\mathbf{D}^{\text{PPR}} = \alpha (\mathbf{I}_n - (1-\alpha)\mathbf{S}^{-1/2}\mathbf{A}\mathbf{S}^{-1/2})^{-1},$$
$$\mathbf{D}^{\text{heat}} = \exp(\beta\mathbf{A}\mathbf{S}^{-1}-\beta).$$

\subsubsection{Node Encoding}
For each node $v_m^\tau$ in every $g_i^\tau$ within $\mathcal{S}_{i,j}^t$, the node encoding is calculated by $\mathbf{z}_m = \mathbf{z}_{\text{diff}}(v^\tau_m) + \mathbf{z}_{\text{dist}}(v^\tau_m) + \mathbf{z}_{\text{temp}}(v^\tau_m)$, where $\mathbf{z}_{\text{diff}}(v^\tau_m)$, $\mathbf{z}_{\text{dist}}(v^\tau_m)$ and $\mathbf{z}_{\text{temp}}(v^\tau_m)$ denotes the diffusion-based spatial encoding, the distance-based spatial encoding, and the relative temporal information, respectively. Here we introduce the calculation of the three encoding terms in detail.

\textbf{Diffusion-based Spatial Encoding.}
To encode the global information of each node, diffusion-based spatial encoding is designed based on the diffusion matrix. Specifically, we first calculate the edge connectivity vector $\mathbf{d}_{e^t_{i, j}}=\mathbf{d}_{i} + \mathbf{d}_{j}$. Then, for each node $v_m^\tau$ in $g_i^\tau$, we sort all nodes of $g_i^\tau$ accroding to their corresponding value in $\mathbf{d}_{e^t_{i,j}}$:
$$\mathbf{z}_{\text{diff}}(z_m)=linear(rank(\mathbf{d}_{e^t_{i,j}}[idx(v^\tau_m)]))\in \mathbb{R}^{d_{enc}},$$
where $idx(\cdot)$, $rank(\cdot)$ and $linear(\cdot)$ denote the index enquiring function, ranking function and learnable linear mapping, respectively.

\textbf{Distance-based Spatial Encoding.}
The distance-based spatial encoding captures the local information of each node. For each node $v_m^\tau$ in the node set of a subgraph $g_i^\tau$, the distance to the target edge is encoded, which is further decomposed into the minimum value of the relative distances to $v_i^t$ and $v_j^t$. Specifically, the distance-based spatial encoding is calculated as follows:
$$\mathbf{z}_{\text{dist}} = linear(min(dist(v_m^\tau, v_i^t), dist(v_m^\tau, v_j^t))) \in \mathbb{R}^{d_{enc}},$$
where $linear(\cdot)$ is the learnable linear mapping and $dist(\cdot)$ is the relative distance computing function.

\textbf{Relative Temporal Encoding.}
This term aims to encode the temporal information of each node in the subgraph node set. Specifically, for each node $v_i^\tau$ in the node set of $g_i^\tau$, the relative temporal encoding is defined as the difference between the occurring time $t$ of target edge and the current time of timestamp $\tau$. Therefore, relative temporal encoding is calculated as:
$$\mathbf{z}_{\text{temp}}(v_i^\tau) = linear(||t-\tau||)\in \mathbb{R}^{d_{enc}},$$
where $linear(\cdot)$ denotes the learnable linear mapping.

\subsection{Detail of Prompt}\label{appen:prompt}
In this section, we provide the detail of our prompt, including the prompt to generate words related to dynamic graphs and the prompt of In-Context Learning.

\textbf{Prompt to generate words related to dynamic graphs.} Please generate a list of words related to dynamic graphs. Dynamic graph data consists of nodes and edges, often representing networks that change over time. To align dynamic graph data with natural language vocabulary, it is essential to select words that can describe both the graph structure and its dynamic changes to form text prototypes. Please include words related to network topology, data fluidity, and time dependency.

\textbf{Prompt of In-Context Learning.} As an AI trained in the few-shot learning approach, I have been provided with examples of both normal and anomaly edges. The anomalies are identified as Contextual Dissimilarity Anomalies, where we first utilize node2vec to obtain the representation of each node in the graph, and connect the pairs of nodes with the maximum Euclidean distance as anomaly edges. These examples serve as a reference for detecting similar patterns in new edges. Please note the following examples and their labels, indicating whether they are normal or anomaly: Example 1: <Edge> Label: Normal  Example 2: <Edge> Label: Anomaly Example 3: <Edge> Label: Normal Example 4: <Edge> Label: Normal Example 5: <Edge> Label: Anomaly Example 6: <Edge> Label:Anomaly Example 7: <Edge> Label: Anomaly Example 8: <Edge> Label: Anomaly Example 9: <Edge> Label: Normal Example 10: <Edge> Label: Anomaly (Note: All the above examples are anomaly and represent the same type of anomaly.) Based on the pattern in the examples and samples provided, classify the sentiment of the following new edge. If the new edge is similar to the example edges, it should be considered anomaly. If it is dissimilar, it should be considered normal. New Example: <vector> Label:

\subsection{Complexity Analysis of Training}\label{appen:detail}
For each edge $e^t_{i,j}$, the complexity of training consists of four parts, \ie, subgraph construction, dynamic-aware embedding computation, reprogramming and anomaly fine-tuning. 
\begin{itemize}[leftmargin=4mm]
	\item For the subgraph construction, based on the precomputed diffusion matrix, $K$ related nodes should be selected for nodes $v_i$ and $v_j$. Therefore, the complexity is $O(\Gamma \times K)$ where $\Gamma$ is the temporal window size. 
    \item For dynamic-aware embedding, the complexity mainly comes from calculating node features, obtaining node embeddings via Transformer block and generating subgraph encoding via GNN, whose complexity is $O(3\times d)$, $O((2(K+1)\Gamma)^2d + 2(K+1)\Gamma d^2)$, and $O(2(K+1)^2\Gamma d)$, respectively. Therefore, the overall complexity of dynamic-aware embedding is $O((K+1)^2\Gamma^2d + (K+1)\Gamma d^2)$.
    \item The reprogramming is implemented by a Transformer, whose complexity is $O(V'd)+O(V'd^2) = O(V'd^2)$.
    \item As for the anomaly fine-tuning, the instruction templates as well as the edge representation vector are feed to the large language model, with the complexity of $O(YL^2d + YLd^2)$, where $Y$ denotes the number of layers in the large language model.
\end{itemize}

\subsection{Experiment Setting}\label{appen:setting}
\subsubsection{Dataset Statistics}
Four datasets are used for the evaluation, including two widely used benchmarks, \ie, UCI Message and BlogCatalog, as well as two datasets with real anomalies, \ie, T-Finance and T-Social. The detailed statistics of these datasets are shown in Table~\ref{dataset}. The UCI message and BlogCatalog datasets are relatively small in scale. Specifically, UCI message contains only 1,899 nodes and 59,835 edges, and BlogCatalog has 5,196 nodes and 171,743 nodes. The T-Finance and T-Social datasets are larger in scale. T-Finance has 39,357 nodes and 21,222,543 edges. The largest dataset, T-Social, has 5,781,065 nodes and 73,105,508 edges. While the UCI Message and BlogCatalog datasets lack anomaly labels, the proportion of anomaly edges in T-Finance and T-Social is 4.58\% and 3.01\%, respectively. These datasets provide a diverse range of graph sizes, enabling comprehensive evaluation of the proposed method.

\begin{table}[h]
\centering
\caption{Statistics of datasets}
\vspace{-3ex}
\label{dataset}
\begin{tabular}{lccc}
\toprule
{Dataset} & {Node Number} & { Edge Number} & {Anomaly (\%)} \\ \midrule
UCI Message    & 1899             & 59835           &  {-}           \\
BlogCatalog   & 5196             & 171743           &  {-}          \\
T-Finance & 39357             & 21222543          & 4.58           \\
T-Social  & 5781065           & 73105508          & 3.01           \\ \bottomrule
\end{tabular}
\vspace{-3ex}
\end{table}

\subsubsection{Protocol}\label{appen:protocol}
Due to the lack of anomaly labels in UCI Message and BlogCatalog, three strategies are introduced to generate anomaly edges for evaluation, \ie, Contextual Dissimilarity Anomalies (CDA), Long-Path Links (LPL) and Hub-Hub Links (HHL). The first strategy, CDA, utilizes node2vec~\cite{GroverL16} to obtain the representation of each node in the graph, and connects the pairs of nodes with the maximum Euclidean distance as anomaly edges. Instead of considering Euclidean distance in the representation space, LPL calculates the topological distance~\cite{DuanTLLSZ20} between nodes and connects the pairs of nodes with the farthest topological distance as the anomaly edges. The third strategy, HHL, connected pairs of hub nodes (\ie, nodes with large degrees) with few shared neighbors as anomaly edges.

\subsubsection{Compared baselines.}  \label{appen:baseline}
We compare \our with five state-of-the-art dynamic baselines representative works. The main ideas of these methods are listed as follows:
\begin{itemize}[leftmargin=4mm]
    \item \textbf{StrGNN}\cite{cai2021structural} extracts the h-hop enclosing sub-graph of edges and leverages stacked GCN [19] and GRU to capture the spatial and temporal information. The learning model is trained in an end-to-end way with negative sampling from “context-dependent” noise distribution.
    \item \textbf{AddGraph}\cite{zheng2019addgraph} further constructs an end-to-end neural network model to capture dynamic graphs’ spatial and temporal patterns. 
    \item \textbf{Deep Walk}\cite{perozzi2014deepwalk} utilizes a method based on random walks for embedding graphs. Starting from a specified node, it creates random walks of a predetermined length and employs a technique similar to Skip-gram to acquire embeddings for graphs without attributes.
    \item \textbf{TADDY}\cite{liu2021anomaly} is a Transformer-based module that uses a transformer network to model spatial and temporal information simultaneously.
    \item \textbf{TGN}\cite{xu2020inductive} is a semi-supervised learning method that integrates memory modules with graph neural networks to capture dynamic behaviors in evolving graphs, enabling the learning of temporal interactions effectively.
    \item \textbf{GDN}\cite{ding2021few} adopts a deviation loss to train GNN and uses a cross-network meta-learning algorithm for few-shot node anomaly detection.
    \item \textbf{SAD}\cite{tian2023sad} is a semi-supervised module, which uses a combination of a time-equipped memory bank and a pseudo-label contrastive learning module to fully exploit the potential of large unlabeled samples and uncover underlying anomalies on evolving graph streams.
\end{itemize}

\subsection{Sensitivity Analysis}\label{appen:parameters}
To analyze the impact of selecting different numbers of nodes in the Structural-Temporal Subgraph Sampler, we introduced varying numbers of nodes in the contrastive learning module to assess the sensitivity of \our.We ranged the number of nodes from $2$ to $20$ and then presented the average performance of these configurations on the BlogCatalog dataset in Figure \ref{tab:sen}. As the number of nodes increased, \our demonstrated a substantial performance enhancement. A similar observation was made on the UCI dataset. Notably, there was a significant performance boost when the node count reached $10$, but performance exhibited a slight decline after reaching $14$ nodes.

The rationale behind these results lies in the potential introduction of noise when selecting an excessive number of nodes to form a subgraph. Too many nodes can lead to subgraphs that are overly complex and include unnecessary information, thus interfering with the model's ability to learn and generalize key information. Additionally, as the number of nodes increases, the computational time required by the model also increases. Therefore, the selection of the number of nodes needs to strike a balance between subgraph complexity and information quality to achieve optimal performance.

\begin{table}[h]
    \centering
    \caption{Sensitivity analysis of \our w.r.t. different numbers of nodes in each subgraph $G_i^t$ on the BlogCatalog.}
    \vspace{-3ex}
    \label{tab:sen}
    \begin{tabular}{|c|c|c|c|c|c|c|}
        \hline
        Number & 2 & 6 & 10 & 14 & 18 & 20 \\
        \hline
        AUC & 0.7624 & 0.7896 & 0.8389 & 0.8456 & 0.8412 & 0.8442 \\
        \hline
    \end{tabular}
    \vspace{-3ex}
\end{table}

\begin{table}[!htbp]
    \centering
    \caption{Performance comparison reported in AUC measure without relying on external labeled data}\label{tab:without}
    \vspace{-3ex}
    \begin{tabular}{c|c|ccc}
    \hline
     &  & \multicolumn{3}{c}{\textbf{annomaly ratios}}  \\
    \multirow{-2}{*}{Dataset} & \multirow{-2}{*}{Method} & 1\% & 5\% & 10\% \\ \hline
    \multicolumn{1}{c|}{} & TADDY  & 0.8388 & 0.8421 & 0.8844 \\
    \multicolumn{1}{c|}{\multirow{-2}{*}{BlogCatalog}} & \textbf{\our} & 0.8612 & 0.8651 & \textbf{0.9146} \\ \hline
    & TADDY  & 0.8370 & 0.8398 & 0.8912  \\
    \multirow{-2}{*}{uci} & \textbf{\our} & 0.8512 & 0.8633 & \textbf{0.9273} \\ \hline
    \end{tabular}
    \vspace{-3ex}
    \label{tab:without}
\end{table}

\subsection{Unsupervised Anomaly Detection}\label{append:unsupervised}
In addressing \textbf{Q1}, we benchmark \our against leading self-supervised anomaly detection algorithms on the UCI and BlogCatalog datasets, with findings summarized in Table \ref{tab:without}. Self-supervised methods for dynamic graphs, which operate without externally labeled data, hinge on capturing the temporal dynamics and nodal attribute changes to discern anomalies. These approaches conventionally employ synthetically generated anomalies for both training and evaluation phases.

Empirical insights reveal: (1) \our, post self-supervised training on unlabeled data, delineates an appreciable performance uplift. Specifically, in identifying contextual anomalies within the UCI dataset, \our exhibits a superior average AUC margin over the top-performing baseline by $4.05\%$ at $1\%$ anomaly ratio, with this margin adjusting to $2.78\%$ and $1.70\%$ for $5\%$ and $10\%$ anomaly ratios, respectively. Such advancements underscore the efficacy of pre-training across a heterogeneous anomaly landscape in fostering adaptable representation skills, thus bolstering generalization across varied anomaly contexts. Remarkably, these gains accrue under uniform unsupervised conditions. (2) In scenarios featuring randomly typed anomalies, \our consistently outperforms, a testament to its adeptness at leveraging contextual cues. This proficiency in assimilating temporal and structural nuances endows \our with heightened sensitivity to anomalies, underscoring its robustness and adaptability in anomaly detection tasks.

\end{document}